\documentclass[11pt]{article}
\usepackage[finalnew]{trackchanges}
\usepackage{multirow}
\usepackage{multicol}
\usepackage{geometry}
\usepackage{setspace}
\usepackage[utf8]{inputenc}
\usepackage{floatrow}
\usepackage{graphicx}
\graphicspath{ {./images/} }
\usepackage{amssymb}
\usepackage{amsmath}
\usepackage{bm}
\usepackage{rotating}
\usepackage{indentfirst}
\usepackage{setspace}
\usepackage[all]{nowidow}
\usepackage[ruled,vlined]{algorithm2e}
\usepackage{titlesec}
\usepackage{tikz}
\usepackage{pdflscape}
\usepackage{rotating}
\usetikzlibrary{shapes.geometric, arrows}
\tikzstyle{startstop} = [rectangle, rounded corners, minimum width=2cm, minimum height=1cm,text centered, draw=black, fill=yellow!20]
\tikzstyle{io} = [trapezium, trapezium left angle=70, trapezium right angle=110, minimum width=0.5cm, minimum height=1cm, text centered, text width=3cm, draw=black, fill=purple!20]
\tikzstyle{process} = [rectangle, minimum width=2cm, minimum height=1cm, text centered, text width=3cm, draw=black, fill=blue!20]
\tikzstyle{decision} = [diamond, minimum width=2cm, minimum height=0.1cm, text centered, text width=3cm,draw=black, fill=green!20]
\tikzstyle{arrow} = [ultra thick,->,>=stealth]

    \makeatletter
    \renewcommand\section{\@startsection {section}{1}{\z@}%
                                       {-3.5ex \@plus -1ex \@minus -.2ex}%
                                       {2.3ex \@plus.2ex}%
                                       {\normalfont\fontfamily{phv}\fontsize{16}{19}\bfseries}}
    \renewcommand\subsection{\@startsection{subsection}{2}{\z@}%
                                         {-3.25ex\@plus -1ex \@minus -.2ex}%
                                         {1.5ex \@plus .2ex}%
                                         {\normalfont\fontfamily{phv}\fontsize{14}{17}\bfseries}}
    \renewcommand\subsubsection{\@startsection{subsubsection}{3}{\z@}%
                                        {-3.25ex\@plus -1ex \@minus -.2ex}%
                                         {1.5ex \@plus .2ex}%
                                         {\normalfont\normalsize\fontfamily{phv}\fontsize{14}{17}\selectfont}}
    \makeatother
 
\usepackage{subcaption}
\usepackage[style=ieee, maxnames=2,minnames=1]{biblatex} 
\usepackage{hyperref}
\allowdisplaybreaks  
  
\addeditor{AI}
\addeditor{AI2}
\addeditor{UJI}
\usepackage{authblk}

\title{\vspace{-1.5cm}\bf\Large Dynamic Exploration-Exploitation Trade-Off in Active Learning Regression with Bayesian Hierarchical Modeling}

\author{\small Upala Junaida Islam$^{a,}$\thanks{Corresponding authors: uislam@asu.edu, aiquebal@asu.edu}, Kamran Paynabar$^b$, George Runger$^a$, and Ashif Sikandar Iquebal$^{a,*}$\\ $^a$School of Computing and Augmented Intelligence, Arizona State University, USA\\ $^b$H. Milton Stewart School of Industrial \& Systems Engineering,
Georgia Institute of Technology, USA }

\date{}
\begin{document}

\maketitle
\renewcommand{\thefootnote}{\arabic{footnote}}
\vspace{-1cm}
\begin{abstract}
    Active learning provides a framework to adaptively query the most informative experiments towards learning an unknown black-box function. Various approaches of active learning have been proposed in the literature, however, they either focus on exploration or exploitation in the design space. Methods that do consider exploration-exploitation simultaneously employ fixed or ad-hoc measures to control the trade-off that may not be optimal. In this paper, we develop a Bayesian hierarchical approach, referred as BHEEM, to dynamically balance the exploration-exploitation trade-off as more data points are queried. To sample from the posterior distribution of the trade-off parameter, We subsequently formulate an approximate Bayesian computation approach based on the linear dependence of queried data in the feature space. Simulated and real-world examples show the proposed approach achieves at least $21\%$ and $11\%$ average improvement when compared to pure exploration and exploitation strategies respectively. More importantly, we note that by optimally balancing the trade-off between exploration and exploitation, BHEEM performs better or at least as well as either pure exploration or pure exploitation. 
\end{abstract}

\noindent%


\section{Introduction}

The past decade has witnessed a widespread adoption of machine learning techniques across a range of applications such as design and discovery of novel materials \cite{lookman}, monitoring and control of advanced manufacturing \cite{wuest2016}, and decision-making in healthcare \cite{char2018}. However, a majority of the success stories of machine learning methods have been limited to fitting large experimental and simulated datasets \cite{Liao2020Jun}. With rising concerns about resource availability
and increasing costs of conducting physical experiments and obtaining labeled datasets, there is a need to develop methodologies that will guide experimental efforts towards gathering the most informative experiments.

Active learning provides a framework to address these challenges by allowing the learning algorithm to adaptively select the most informative experiments in learning a concept (a classification or a regression model), reducing the need for obtaining large datasets \cite{settles.survey}.
Here, the informativeness of data is usually assessed via an acquisition function. Following the seminal work of Angluin on ``Queries and Concept Learning'' \cite{angluin1988} where the acquisition function was a majority vote among a set of candidate hypotheses, several acquisition functions were introduced that are based on entropy \cite{holub2008}, prediction uncertainty \cite{cohn1996}, etc. 
However, recent studies have shown that they may have exploration or exploitation bias \cite{Cebron}. For instance, a greedy sampling-based acquisition function proposed in \cite{Wu2019} (see Equation~\eqref{igs}), tends to acquire the data points uniformly in the unexplored search space and therefore favors exploration, irrespective of the underlying target concept. An example of this is shown in Figure~\ref{toyproblem}(a). In contrast, a Query-by-Committee (QBC) approach for active learning \cite{Burbidge} tends to exploit the search space in the neighborhood where the target concept violates the continuity and, in some cases, uniform continuity assumptions. An example is shown in Figure~\ref{toyproblem}(b).

This raises the classic, albeit important question: how can we encode exploration-exploitation trade-off in active learning for regression problems. Based on our survey of the literature, we note two prominent schools of thought. The first is based on constructing new acquisition functions or appropriate modifications thereof to balance exploration and exploitation \cite{loy2012stream}. An example of the latter is the hierarchal expected improvement \cite{Chen2021} that was introduced in an attempt of achieving non-myopic search in the context of Bayesian optimization. This was accomplished by modifying expected improvement \cite{Jones1998Dec} to encourage more exploration. But the most commonly investigated approach is based on optimizing a combination of different acquisition functions for exploration and exploitation through a trade-off parameter 
\cite{Cebron} as, 

\vspace{-1em}
\begin{equation}
\textbf{x}^*=\underset{\mathbf{x}}{\arg\!\max}\left(\eta \mathcal{F}_1(\mathbf{x})+(1-\eta) \mathcal{F}_2(\mathbf{x})\right)\label{eq:1}
\end{equation}
\vspace{-3em}

\noindent Here, $\eta\in[0,1]$ is a parameter controlling the trade-off between $\mathcal{F}_1(\cdot)$ and $\mathcal{F}_2(\cdot)$, the objective functions corresponding to exploration and exploitation acquisition functions respectively, and $\mathbf{x}^*$ is the optimal data to query according to the new strategy. Nonetheless, the challenge with this approach is that $\eta$ is unknown and needs to be estimated during the learning process. Since it is difficult to estimate $\eta$ without observing future data points, existing approaches have relied on trial and error \cite{Ajdari} or predefined ad-hoc measures depending on the number of queried data \cite{Elreedy}. The exploration-exploitation bias has received attention in the reinforcement learning literature as well. 
However, unlike active learning, the agent in reinforcement learning is only concerned with maximizing the long-term reward by finding an optimal sequence of actions via exploration-exploitation. 
Conversely, in active learning, we are concerned with maximizing the reward over a finite (often very small) number of experiments that we can perform. The active learner, therefore, needs to iteratively and effectively update the trade-off parameter as new data are queried.

During active learning, the trade-off parameter is influenced by the noise in the labeled data at each query stage and by the variability in the labeled data across different query stages. This bi-level structure allows us to use a Bayesian hierarchical model to find the optimal trade-off parameter during the learning process. Here, the dependency of parameters related to the hierarchy is reflected in a joint probability model. The posterior density and uncertainty quantification of the parameters are obtained via Bayesian inference \cite{gelman1995}. 
In the absence of closed-form expressions for the posterior distribution, a sampling-based Markov chain Monte Carlo (MCMC) method is used to obtain the marginal posterior distributions~\cite{guo2009bayesian}.

To this end, we present a Bayesian Hierarchical Model to automatically balancing Exploration-Exploitation (referred as BHEEM) trade-off in actively learning an unknown black box function. 
In particular, BHEEM
iteratively updates the trade-off parameter as data points are queried to minimize the generalization error. By imposing a hierarchical structure, we account for the variability in $\eta$ arising from within each query iteration as well as across subsequent queries. We present an Approximate Bayesian Computation (ABC) approach along with Metropolis within Gibbs algorithm to sample from the posterior distribution of $\eta$. We subscribe to Gaussian Process Regression to obtain the best fit for the underlying black-box function. It is critical to note that the proposed methodology is generalizable and not contingent on the choice of exploration and exploitation or the choice of the regression function. We evaluate the efficacy of BHEEM on six simulated and benchmark datasets, and one real-world example in materials discovery. We also provide sensitivity analysis and numerical convergence results to establish the consistency of BHEEM.

The rest of the paper is organized as follows. Section 2 presents a brief relevant literature review. Section 3.1 unfolds the Bayesian hierarchical model for estimating the trade-off parameter to balance exploration and exploitation. Section 3.2 explains the Gaussian process regression methodology we employed active learning with. Exploration and exploitation approaches adopted in this work are presented in Section 3.3 followed by trade-off strategies commonly used in the literature summarized in Section 3.4. Experimental setup, results from simulation and real-world experiments, and analysis over sensitivity and convergence are presented in Section 4. The paper is concluded in~Section~5.

\section{Literature Review}

The roots of active learning can be traced back to the early 1960s with initial works focused on space-filling designs via sequential experimentation \cite{chernoff1959}. Later studies developed model-based sequential experimental designs where an underlying data-generating model is considered to guide the search for the subsequent experiments \cite{joseph2015sequential,chen2017sequential,lu2021strategies}. Model-based sequential design is commonly referred to as active learning in the machine learning community \cite{settles.survey}. Over the past two decades, active learning has witnessed significant developments in intelligently querying and labeling data, both in the classification \cite{dasgupta2008hierarchical,li2013adaptive,beluch2018power} and regression tasks \cite{Burbidge,Cai,Wu2019}. In this section, we investigate active learning in regression problems with a particular emphasis on the exploration-exploitation~trade-off.

Two pioneering works in active learning for regression problems are Active Learning - Mackay (ALM) based on maximizing the expected information gain \cite{mackay1992} and Active Learning - Cohn (ALC) based on minimizing the generalization error \cite{cohn1996}. 
Mackay defined the information gain as the change of Shannon's entropy \cite{Shannon} before and after labeling data. Entropy has also been formulated as a measure for querying data in classification problems within the framework of ``uncertainty sampling" \cite{lewis1994heterogeneous}. 
Cohn et al. \cite{cohn1996}, on the other hand, demonstrated that the expected generalization error is composed of data noise, model bias, and variance. Data noise is independent of the model, and model bias is invariant given a fixed model. Thus the criteria for querying data was simplified to the minimization of the total variance. Meka et al. \cite{Meka} extended the ALC strategy by adding a regularizing term to integrate the information of both queried and unqueried data.

A more theoretically-motivated active learning strategy called Query-by-Committee (QBC) \cite{Seung92} was first developed for regression by Krogh and Helseby \cite{krogh95}. QBC maintains a committee of hypotheses that are simultaneously trained on the labeled data. The data that maximizes the disagreement between the committee members is deemed the most informative. 
A variation of QBC, 
Expected Model Change Maximization (EMCM) was also developed in a regression setting where the expected model change was constructed using the gradient of the error concerning candidate data \cite{Cai}.
O'Neil et al. \cite{oneill2017} compared QBC and EMCM along with a few model-free strategies that select the unlabeled data based on density, or diversity, and through an acquisition function named Exploration Guided Active Learning (EGAL) combining both density and diversity \cite{hu2010}. 
Results from O'Neal et al. \cite{oneill2017} indicated that the integral properties of the dataset, such as its geometry, can be promising candidates for active learning strategies.
A similar diversity-based passive sampling strategy, Greedy sampling (GS) was adapted as improved Greedy Sampling (iGS) for active learning by incorporating the response information ensuring exploration in both input and output space \cite{Wu2019}. 

While significant developments have taken place in the active learning literature, only a handful of the efforts have considered the problem of exploration-exploitation trade-off, the majority of which is limited to classification problems. For example, past researches have used maximum entropy \cite{holub2008}, and distance and similarity-based metrics for exploration \cite{baram2004} while subscribing to mostly uncertainty and redundancy to exploit near the current decision boundary \cite{yin2017}. Approaches to handle the exploration-exploitation bias include switching between exploration and exploitation based on their performance at each iteration \cite{baram2004}, explore-then exploit approach \cite{smith2020imprecise}, random probabilistic measure \cite{Elreedy}, or combining exploration and exploitation in a predefined ratio \cite{Cebron,oneill2017,kee2018}. Combining the acquisition functions based on cross-validation, sensitivity analysis, or trial-and-error \cite{Ajdari} can only be performed retrospectively and does not help with prospective data~queries.

Beyond active learning, the exploration-exploitation problem has also been investigated in the reinforcement learning literature. Here, the exploitation of the greedy approach has been confronted with various exploratory approaches. Upper Confidence Bound (UCB) is one of the most celebrated approaches against this bias. Here, uncertainty is used to balance the exploration and exploitation instead of choosing the greedy action according to current knowledge \cite{auer2002using}. The $\epsilon-$greedy approach decays the randomness or the rate of exploration over the learning process to encourage more exploitation of the observed knowledge \cite{afifi2020}. The Boltzmann exploration approach uses exponential weighting schemes to balance exploration and exploitation in a probabilistic fashion \cite{sutton2018}. The balancing parameter has also been obtained using a trial-and-error process \cite{koulouriotis2008reinforcement} or controlled based on a variation of action results and perception of environmental change~\cite{ishii2002control}. 

Based on the literature review, we notice that the existing methods have either considered either a predefined ratio, probabilistic function or trial and error methods to control the exploration-exploitation trade-off. None of the efforts have focused on adaptively updating the trade-off parameter as more data is queried. This study advances the research on dynamically updating the trade-off between exploration and exploitation as more knowledge is gathered about the black-box function from the queried data.

\section{Methodology}

\begin{figure}[h!]
    \centering
    \includegraphics[width=0.8\linewidth]{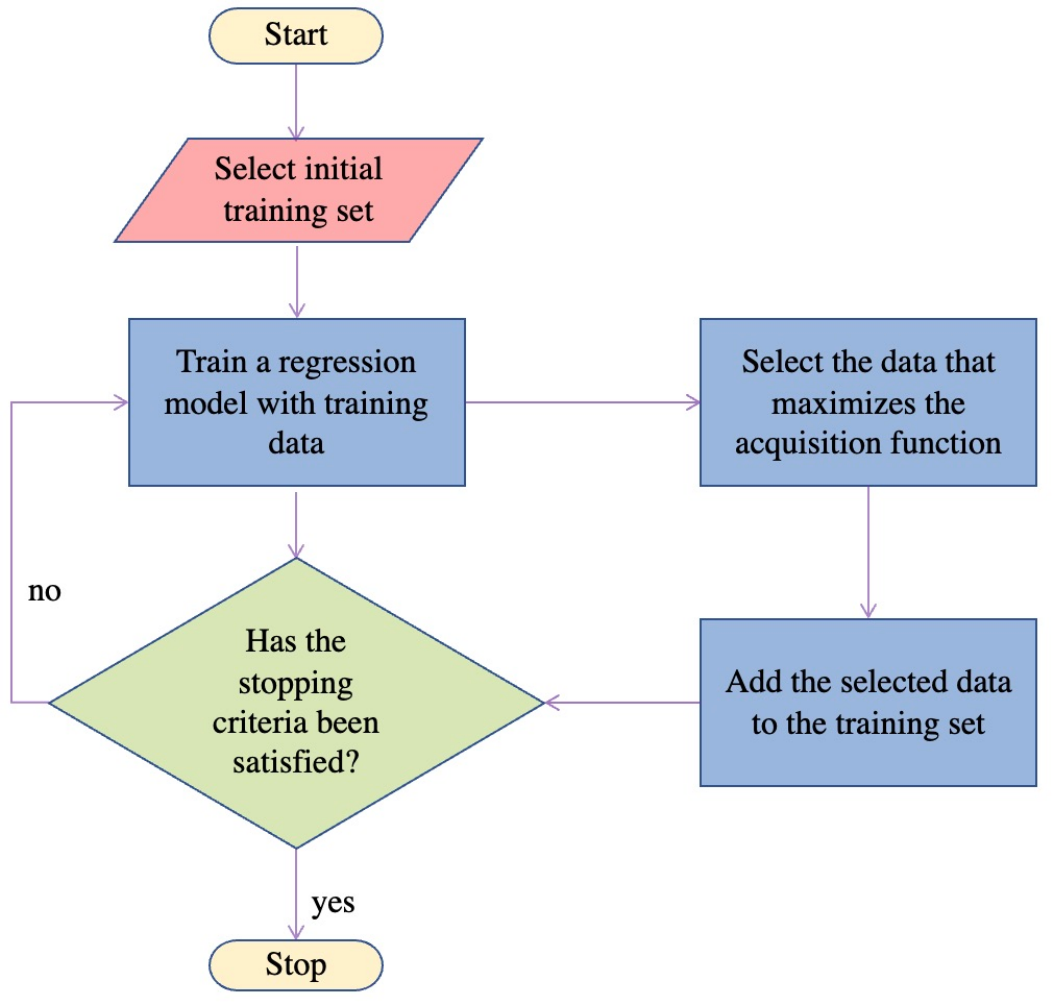}
    \caption{General schema of an active learner.}
    \label{fig:flowchartAL}
\end{figure}

In this section, we present the general schema of active learning along with the models and strategies we employ in this study. 
Let $X=\{\mathbf{x}_1,\mathbf{x}_2,\hdots\}$ and $\mathbf{y}=\{y_1,y_2,\hdots\}$ represent the possible collection of all design points and their corresponding responses respectively. In the generic problem of active learning, we consider that 
the actively growing set of queried (i.e., labeled) data $\mathcal{D}^N=\{(\mathbf{x}_{o_i},y_{o_i})\}_{i=1}^N$ is accessed by the underlying model. Here, $\mathbf{x}_{o_i}{\in}\mathbb{R}^d$ and $y_{o_i}{\in}\mathbb{R}$
are the queried data and their corresponding output responses respectively. The objective of active learning in a regression setting is to select the next data $\mathbf{x}_{o_{N+1}}$ from the search space, $X$, which will be the most informative towards learning the unknown black-box function $f(\mathbf{x})$ given the current knowledge. The response $y_{{o}_{N+1}}$ is obtained via simulation or physical experiments. By actively selecting the data to learn from, it can rapidly reduce the generalization error which is the prediction error of the algorithm at the data unobserved so far. The process is iterated until stopping criteria e.g., the maximum number of labeled data, or predicted change, have been satisfied. 
See Figure~\ref{fig:flowchartAL} for a flow chart of the active learning approach.

\subsection{\emph{Exploration-exploitation trade-off}}

In machine learning, exploration refers to generating novel information in the search space while exploitation focuses on improving decisions (e.g., maximizing reward) based on the available information. Although this is a generally accepted definition of exploration and exploitation \cite{cohenshould}, its usage varies depending on the context. Particularly, in the regression setting, exploration refers to sampling data from the unobserved regions in the search space to gather more (potentially novel) information about the black-box function, such as local minima/maxima. In contrast, exploitation aims at accurately capturing the black-box function in the regions where it is highly unpredictable and has a sharp change or discontinuity. As mentioned in the foregoing, it is possible to encode exploration and exploitation separately in an acquisition function \cite{hu2010}, however, it is challenging to dynamically balance between the two \cite{rabbitt1966}. We consider a general framework for sampling data points that maximizes a linear combination of acquisition functions aimed at simultaneous exploration and exploitation as presented in Equation~\eqref{eq:1}. The next data is queried through pure exploitation when $\eta=0$, and pure exploration when $\eta=1$. By increasing $\eta$ from $0$ to $1$, the degree of exploration increases in the constructed acquisition function.

To estimate $\eta$ dynamically, we consider two levels of variability. First is the variability in queried data emerging from the noise and measurement errors in $y$ at each querying stage. The second level of variability is associated with the dynamic nature of $\eta$ between querying stages due to the nature of the unknown black-box function (e.g., jumps or discontinuities). This bi-level nature of variability motivates the need for a hierarchical model. A Bayesian hierarchical model allows us to capture this hierarchy while simultaneously encoding the uncertainty at each of the levels. The trade-off parameter $\eta_j$ captures the first level of variability at querying stage $j$ given the set of hyper-parameters $\bm{\theta}$. We model $\{\eta_j|\bm{\theta}_j\}\stackrel{iid}{\sim} p(\eta|{\bm{\theta}_j})$. Here, the conditional independence of $\{\eta_j|\bm{\theta}_j\}$ tells us that the trade-off parameters at one sampling stage are exchangeable (de Finetti's theorem \cite{bernardo1996}), but not independent.
To capture the variability in $\eta$ across multiple sampling stages, we model $\{\bm{\theta}_j|\bm{\psi}\}\stackrel{iid}{\sim} p(\bm{\theta}|\bm{\psi})$ where $\bm{\psi}$ is another set of  hyper-parameters fixed from \emph{a priori} assumptions. De Finetti's theorem from this conditional independence shows that the trade-off parameters at subsequent querying stages are exchangeable as well. Indeed, this aligns with our assumption that the trade-off parameter at any querying stage is not completely independent of the previous actions (exploration or exploitation). 
In the next two subsections, we discuss the prior and the sampling distributions employed in this study.

\subsubsection{\emph{Prior distribution}}
\begin{figure}[h]
     \centering
     \includegraphics[width=0.6\linewidth]{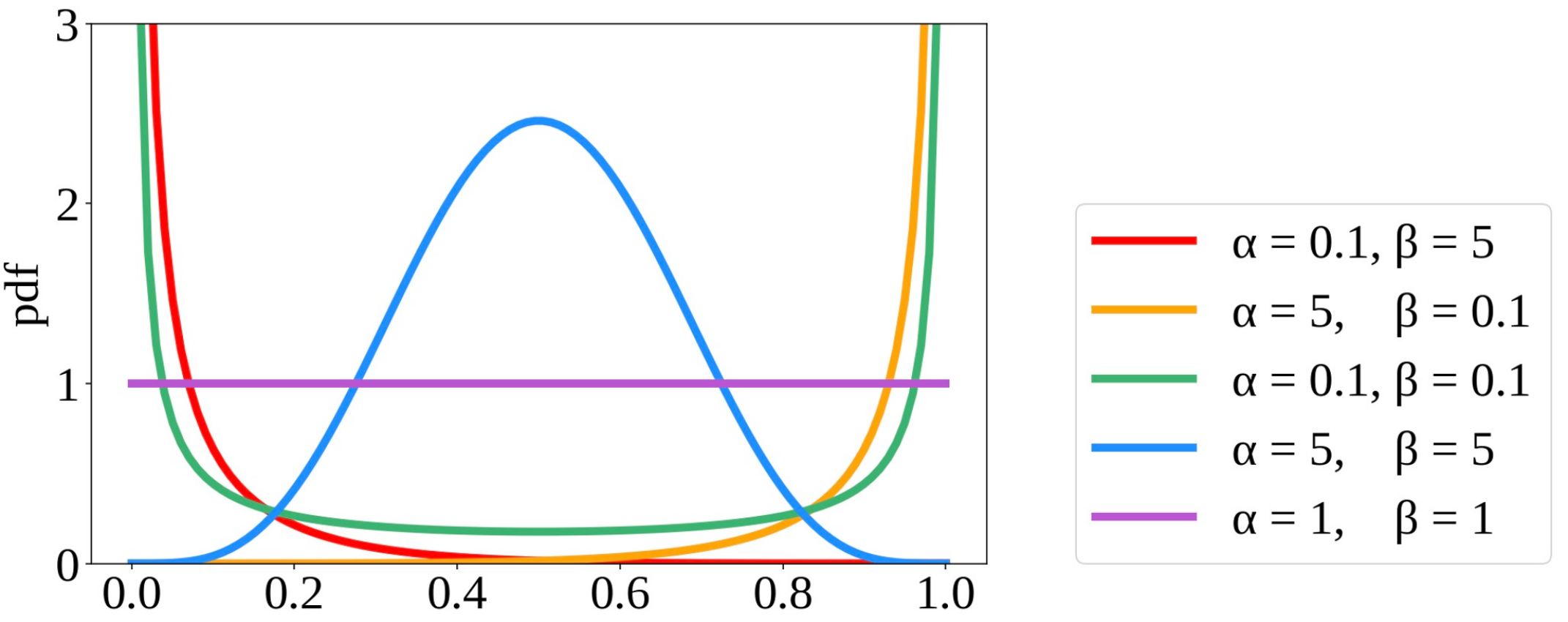}
     \caption{Probability density function for the beta distribution.}
     \label{betapdf}
 \end{figure}

Towards optimally estimating $\eta$ using a Bayesian hierarchical framework, we begin by considering a prior distribution based on the knowledge that it lies between $0$ (pure exploitation) and $1$ (pure exploration). With this knowledge, we let $\eta$ follow a beta distribution in the form of $\eta\sim\text{Beta}(\alpha,\beta)$ with hyperparameters $\alpha$ and $\beta$. The combined selection of $\alpha$ and $\beta$ defines the shape of the prior distribution of $\eta$ as observed in Figure~\ref{betapdf}.
A larger value of $\alpha$ shifts the bulk of the probability towards 1 and emphasizes exploration (e.g., $\alpha=5,\beta=0.1$), whereas an increase in $\beta$ moves the distributions towards 0 and encourages exploitation (e.g., $\alpha=0.1,\beta=5$). $\alpha=\beta<1$ creates a U-shape distribution with maximum probability near 0 and 1 imposing a similar higher probability on both exploration and exploitation (e.g., $\alpha=\beta=0.1$). On the other hand, $\alpha=\beta>1$ generates a bell-shape distribution with the maximum probability at the middle $\left(\eta=0.5\right)$ imposing a lower probability on pure exploration and exploitation, and a higher probability on a mixture of them (e.g., $\alpha=\beta=0.5$). Finally, $\alpha=\beta=1$ creates a uniform distribution with equal probability everywhere and is indifferent towards the selection of $\eta$. Based on extensive simulations, we note that changing the range of $\alpha$ and $\beta$ beyond the interval of $[a,b]=[0.1,5.0]$ has no significant impact on the exploration-exploitation trade-off. Without any information about the nature of the black-box function, we impose a uniform prior distribution in the form of $\alpha,\beta\sim \text{Uniform}(0.1,5.0)$.

\subsubsection{\emph{Posterior and sampling distribution}}
The joint posterior distribution of $\eta$ and the hyper-parameters is,

\vspace{-1em}
\begin{equation}
    p(\eta,\alpha,\beta|\mathbf{y},X)\propto p(\alpha)p(\beta)p(\eta|\alpha,\beta)p(\mathbf{y},X|\eta)\label{posterior} 
\end{equation}
\vspace{-3em}

\noindent where $\{\left(X,\mathbf{y}\right)\}$ represents the collection of both labeled and unlabeled data. In Equation~\eqref{posterior}, $p(\mathbf{y},X|\eta)$ is the likelihood of $\eta$ denoted by $\mathcal{L}(\eta;\mathbf{y},X)$, and $p(\eta|\alpha,\beta)$ is the prior. The hyperprior $p(\alpha)p(\beta)=p(\alpha|a)p(\beta|b)$ due to the independence of $\alpha$ and $\beta$ as well as \emph{a priori} selection of $a$ and $b$. The joint posterior distribution in Equation~\eqref{posterior} is not available in the closed form due to the intractable likelihood function, $p(\mathbf{y},X|\eta)$. Therefore, the unknown parameters are sampled from their respective full conditional distributions using Gibbs sampling. The individual full conditional distribution of $\alpha, \beta, \eta$ is factorized as,
\vspace{-1em}
\begin{align}
      &{}p(\alpha|\beta,\eta)\propto{p(\alpha)}{p(\eta|\alpha,\beta)}\nonumber\\
      &p(\beta|\alpha,\eta)\propto{p(\beta)}{p(\eta|\alpha,\beta)}\nonumber\\
      &p(\eta|\alpha,\beta,\mathbf{y},X)\propto{p(\eta|\alpha,\beta)}{p(\mathbf{y},X|\eta)}\label{full-cond}
\end{align}
\vspace{-3em}

The Gibbs sampler proceeds by iteratively constructing a dependent sequence of parameter values whose distribution converges to the target joint posterior distribution \cite{hoff2009}. But  
the likelihood of $\eta$ is intractable unlike that of $\alpha$ and $\beta$ since the generation of new data from observed $\eta$ does not follow any known distribution. In the absence of a tractable likelihood, we subscribe to Approximate Bayesian Computation (ABC) to approximate the posterior distribution of $\eta$ which has been used in numerous previous applications where evaluation of the likelihood is expensive or infeasible \cite{busetto2009}.

The traditional ABC algorithm \cite{pritchard1999population}, also known as ABC rejection sampler, follows a two-step procedure to sample from the posterior distribution. First, it draws samples of the unknown parameter from its prior distribution. Second, the sampled parameter values are accepted based on the similarity between the data simulated under some model specified by the sampled parameter values and the queried data, or between the summary statistics thereof. In other words, ABC uses summary statistics to filter samples that do not agree with the queried data. Though efficient in simple probability models, ABC rejection is limited in complex ones where the posterior is greatly different from the prior since it involves sampling from the prior \cite{marjoram2003markov}. Different computational methods have been presented in the literature to overcome this limitation of ABC inference, among which ABC-MCMC and its variants have been widely used and accepted \cite{marjoram2003markov}. The ABC-MCMC algorithm combines ABC with Metropolis-Hastings or Metropolis algorithm for iterative approximate Bayesian computation. In the following, we will discuss the Metropolis sampling followed by a new approach to implement ABC and summary statistics into BHEEM.

Metropolis sampling is a simplified form of the Metropolis-Hastings sampling algorithm. It proposes a symmetric proposal distribution $J(\mathbf{\theta}|\mathbf{\theta}_{old})$ for the parameter vector $\mathbf{\theta}$ given the most recent accepted sample $\mathbf{\theta}_{old}$ at each sampling stage. Each proposed sample is either accepted or rejected depending on an acceptance ratio computed from the prior and likelihood of the proposed and old samples. 
ABC-MCMC algorithm shadows the steps of a typical Metropolis algorithm. The initial sample is drawn from the prior, and the next ones are sampled from a proposal distribution centered at the most recently accepted sample. But the acceptance ratio constructed by ABC uses summary statistics, therefore replacing the likelihood ratio like in the regular Metropolis algorithm.

We initiate the Gibbs sampler with $\bm{\mathbf{\theta}}^{(0)}=\{\alpha^{(0)},\beta^{(0)},\eta^{(0)}\}$, and generate a dependent sequence of the parameter vector, $\{\bm{\mathbf{\theta}}^{(1)},\bm{\mathbf{\theta}}^{(2)},\hdots,\bm{\mathbf{\theta}}^{(s)}\}$. Given a current state of the parameters $\bm{\mathbf{\theta}}^{(s)}=\{\alpha^{(s)},\beta^{(s)},\eta^{(s)}\}$, the next state of the parameter vector is generated as follows,
 
\textbf{Step 1:} Sample $\alpha^{(s+1)}\sim p(\alpha|\beta^{(s)},\eta^{(s)})$

\textbf{Step 2:} Sample $\beta^{(s+1)}\sim p(\beta|\alpha^{(s+1)},\eta^{(s)})$

\textbf{Step 3:} Sample $\eta^{(s+1)}\sim p(\eta|\alpha^{(s+1)},\beta^{(s+1)},\mathbf{y},X)$: As mentioned before, we employ ABC-MCMC at this stage. Similar to a typical Metropolis algorithm, we sample $\eta^{(s+1)}$ till it converges. We propose $\hat{\eta}$ from a proposal symmetric distribution centered at $\eta_{old}^{(s+1)}$. The proposed sample is accepted $\left(\eta_{new}^{(s+1)}=\hat{\eta}\right)$ with probability $\text{min}(1,r_{\eta})$ or rejected $\left(\eta_{new}^{(s+1)}=\eta_{old}^{(s+1)}\right)$ with probability $1-r_{\eta}$ where the acceptance ratio, $r_{\eta}$, is constructed as, 
\vspace{-1em}
\begin{align*}
    r_{\eta} = \frac{p\left(\hat{\eta}|\alpha,\beta,\mathbf{y},X\right)}{p\left(\eta_{old}^{(s+1)}|\alpha,\beta,\mathbf{y},X\right)}=\frac{p\left(\hat{\eta}|\alpha_{new}^{(s+1)},\beta_{new}^{(s+1)}\right)p\left(\mathbf{y},X|\hat{\eta}\right)}{p\left(\eta_{old}^{(s+1)}|\alpha_{new}^{(s+1)},\beta_{new}^{(s+1)}\right)p\left(\mathbf{y},X|\eta_{old}^{(s+1)}\right)}
\end{align*}
\vspace{-3em}

But we do not have access to an explicit expression for the likelihood $p(\mathbf{y},X|\eta)$. To check the fitness of the proposed sample $\hat{\eta}$ and to compare it with the current sample $\eta_{old}^{(s+1)}$ in the absence of the likelihood, ABC reconstructs the acceptance ratio $r_{\eta}$ as,
\vspace{-1em}
\begin{align}
    r_{\eta}=\frac{p\left(\hat{\eta}|\alpha_{new}^{(s+1)},\beta_{new}^{(s+1)}\right)}{p\left({\eta}_{old}^{(s+1)}|\alpha_{new}^{(s+1)},\beta_{new}^{(s+1)}\right)}\mathbb{I}\{d(\mathcal{S}(\hat{\mathbf{x}}),\mathcal{S}(X_o))\geq\nu\}
\end{align}
\vspace{-3em}

\noindent where $\mathcal{S}(\hat{\mathbf{x}})$ and $\mathcal{S}(X_o)$ are the sufficient statistics for data nominated by $\hat{\eta}$ and the queried data respectively, $d(\cdot,\cdot)$ is a distance measure between the two statistics, and $\mathbb{I}\{\cdot\}$ is an indicator function. In traditional ABC, the fitness of a candidate sample is measured in terms of its similarity to the observed data. However, in active learning, it is measured in terms of the new information generated by each new query.  
To measure the fitness of candidate queries, we utilize their linear dependency on the already queried data in the Hilbert space. Any data $\hat{\mathbf{x}}$ is considered approximately linearly dependent (ALD) on the already queried set $X_o$ if, 
\vspace{-1em}
\begin{align}
    \delta=\underset{\mathbf{a}}{\min}{\left\Vert{\sum_{i=1}^{N-1}a_i\mathbf{\phi}(\mathbf{x}_{o_i})-\mathbf{\phi}(\hat{\mathbf{x}})}\right\Vert}^2\leq\nu\label{delta}
\end{align}
\vspace{-3em}

\noindent $\nu$ is an threshold parameter determining the level of sparsity, and $\phi$ is a finite-dimensional mapping of the feature vectors, $\mathbf{x}$, to a Hilbert space. If the ALD condition in Equation~\eqref{delta} holds, 
$\phi(\hat{\mathbf{x}})$ can be inferred by the already queried data, $X_o$, as shown in the expression that $\phi(\hat{\mathbf{x}}) = \sum_i a_i \phi(\mathbf{x}_{o_i})$. In this case, no new information is gained and we reject the candidate query. In contrast, if the ALD criterion is not satisfied, i.e., $\delta >\nu$, then we select the candidate query and collect its response.  

However, to use the ALD criterion in order to measure the fitness of candidate queries, we need to define the mapping $\phi$. In the absence of the function $phi$, we invoke Mercer's Theorem \cite{mercer1909xvi} that guarantees the existence of a kernel function $k(\mathbf{x}, \mathbf{x}')$ such that $k= \langle\phi(\mathbf{x}), \phi(\mathbf{x}')\rangle$. As a result, all the calculations in the feature space can be performed by defining the kernel function instead of the function $\phi$. With the substitution of $\phi$ with the kernel function, we obtain, 
\vspace{-1em}
\begin{align}
    \mathbf{a}^*=K(X_o,X_o)^{-1}K(X_o,\hat{\mathbf{x}})\quad\text{ and }\quad \delta =K(\hat{\mathbf{x}},\hat{\mathbf{x}})- K(X_o,\hat{\mathbf{x}})^T\mathbf{a}^*
\end{align}
\vspace{-3em}

For each candidate $\hat{\eta}$ proposed by the Metropolis algorithm, we check the linear dependency of the nominated query $\hat{\mathbf{x}}$. In particular, $\hat{\mathbf{x}}$ is accepted for querying if $\delta \geq \nu$ such that $\mathbb{I}\{\delta \geq \nu\} =1$. In that case, we compare the prior of $\hat{\eta}$ and $\eta_{old}^{(s+1)}$ before accepting one of them as $\eta_{new}^{(s+1)}$. On the other hand, if $\delta<\nu$, it makes $\mathbb{I}\{\delta\geq\nu\}=0$, implying $\eta_{new}^{(s+1)}=\eta_{old}^{(s+1)}$ irrespective of the priors. Authors in \cite{engel2004kernel} experimented with different values for $\nu$ using cross-validation. Through the sensitivity analysis $\nu$ as discussed in Section 4.5, we fixed $\nu=0.001$ for all our experimentation. 

\textbf{Step 4:} Let $\bm{\theta}^{(s+1)} = \{ \alpha^{(s+1)}, \beta^{(s+1)}, \eta^{(s+1)}\}$. After generating  $\{\bm{\theta}^{(1)},\bm{\theta}^{(2)},\hdots,\bm{\theta}^{(s)}\}$, the approximation of $\eta_j$ is obtained by averaging the observed $\{\eta_j^{(1)},\eta_j^{(2)},\hdots,\eta_j^{(s)}\}$.

In this work, we defined the proposal distribution as $\mathcal{N}({\eta}_{old},\tau^2)$ where ${\eta}_{old}$ is the sample parameter accepted in the last iteration, and $\tau^2$ is the variance. The performance of the MCMC chain depends on the variance of proposal distribution, $\tau^2$. If $\tau$ is too small, i.e., 0.001, the Metropolis can get stuck at one sample and may take a long time for the chain to converge. On the other hand, if $\tau$ is too large, i.e., 0.25, it is possible to sample highly diverse and sometimes infeasible values increasing the rejection rate as well as hampering the rate of convergence. A detailed sensitivity analysis for the effect of $\tau$ on the convergence of the algorithm is discussed in Section 4.4.

\subsection{\emph{Gaussian Process Regression}}

In the absence of any known functional form of $f(\mathbf{x})$, we consider Gaussian process regression (GPR) as our underlying learning model. GPR considers a distribution over the underlying function $f(\mathbf{x})$ and aims to specify the black-box function by its mean and covariance. Due to the noise inherited in labeled data without the learner's knowledge, the output $y$ comprises $f(\mathbf{x})$ as    $y = f(\mathbf{x})+\varepsilon$ where $\varepsilon\sim\mathcal{N}\left(\mathbf{0},\sigma_n^2\right)$. Observing the noisy outputs, ${\mathbf{y}}_o$, GPR attempts to reconstruct the underlying function $f(\mathbf{x})$ by removing the contaminating noise, $\varepsilon$ \cite{Rasmussenbook}. We denote the queried dataset as $({X}_o,\mathbf{y}_o)$. GPR imposes a zero-mean Gaussian process prior over the noisy outputs such~that,

\vspace{-1em}
\begin{equation}
  \mathbf{y}_o\sim \mathcal{N}\left(\mathbf{0},K({X}_o,{X}_o)+\sigma_n^2I\right)   
\end{equation}
\vspace{-3em}

\noindent The joint prior distribution between the training output set $\mathbf{y}$ and test output set $\hat{\mathbf{f}}$ is as~following,

\vspace{-1em}
 \begin{equation}    \begin{bmatrix}\mathbf{y}\\\hat{\mathbf{f}}\end{bmatrix}\sim\mathcal{N}_m\left(\mathbf{0},\begin{bmatrix}K(X_o,X_o)+\sigma_n^2I & K(X_o,X_u)\\
    K(X_u,X_o) & K(X_u,X_u)\end{bmatrix}\right)
 \end{equation}
where $X_u$ is the set of unlabeled testing points. The posterior distribution at the test data is given as $\{\hat{\mathbf{f}}|X_o,\mathbf{y}_o,X_u\}\sim\mathcal{N}\left(\bar{\mathbf{f}},\text{cov}(\hat{\mathbf{f}})\right)$~where,
\vspace{-1em}
\begin{align}
    \bar{\mathbf{f}}&{}=K(X_u,X_o)[K(X_o,X_o)+\sigma_n^2I]^{-1}\mathbf{y}_o\label{mpred}\\
    \text{cov}(\hat{\mathbf{f}})&=K(X_u,X_u)-K(X_u,X_o)[K(X_o,X_o)+\sigma_n^2I]^{-1}K(X_o,X_u) \label{vpred}
\end{align}

\subsection{\emph{Acquisition functions }}
In this section, we present some of the most commonly used acquisition functions and characterize them either as an exploration or an exploitation strategy based on the literature and our extensive simulation studies. 

$\bullet\quad \textbf{Improved Greedy Sampling: }$
As mentioned in Section 2, the concept of improved greedy sampling (iGS) was introduced to ensure diversity in the queried data \cite{Wu2019}. At every iteration, iGS determines the unexplored region by searching over the entire region in both the input and output spaces. It queries the next data which is located the farthest from its nearest training point or in an unobserved region according to the then prediction by the regression model.
The acquisition function at $\mathbf{x}_j$ is calculated as $\min\left(u(\mathbf{x}_j)v(\mathbf{x}_j)\right)$ where $u(\mathbf{x}_j)=\|\mathbf{x}_j-X_{o}\|_2$ and $v(\mathbf{x}_j)=\|\bar{f}_j-\mathbf{y}_{o}\|_2$ refer to the distance of $\mathbf{x}_j$ with training points in input and output space respectively. Here $\mathbf{y}_{o}$ is the set of the observed output or the labels at training points $X_{o}$, $\bar{f}_j$ is the predicted output at test point $\mathbf{x}_j$, and $\|\cdot\|_2$ is the L2 norm.
IGS defines the distance at $\mathbf{x}_j$ as $u(\mathbf{x}_j)v(\mathbf{x}_j)$ instead of $u(\mathbf{x}_j)+v(\mathbf{x}_j)$ or $\left(u(\mathbf{x}_j)\right)^2+\left(v(\mathbf{x}_j)\right)^2$ due to the possibility of significantly different scale of $u$ and $v$ which can hamper the latter two formulas with the dominance of one measure over another. It then selects the next data, $\mathbf{x}_{\text{iGS}}^*$, such that,

\vspace{-1em}
\begin{equation}
     \mathbf{x}_{\text{iGS}}^*=\underset{\mathbf{x}_j}{\arg\!\max}{\left(\min{\left(\|\mathbf{x}_j-X_{o}\|_2\|\bar{f}_j-\mathbf{y}_{o}\|_2\right)}\right)}   \label{igs}
\end{equation}
\vspace{-3em}

\noindent By selecting data from unexplored regions in both input and output spaces, iGS avoids sampling from any concentrated region and hence satisfies the requirement for an exploration strategy.

$\bullet\quad \textbf{Query by Committee: }$ Successfully applied in classification and regression-based problems, Query by Committee (QBC) is a framework rooted in the concept of utilizing an ensemble of hypotheses \cite{Seung92}. Maintaining a committee of models, QBC queries the data where the committee members disagree the most about a measure of criteria. The committee members are all trained on the same set of training points but with competing hypotheses or regression models.
Considering a committee of $Q$ models denoted as $h_1,h_2,\ldots,h_Q$, we define the measure of disagreement between two models $h_l$ and $h_p$ at $\mathbf{x}_j$ as the absolute difference of prediction by the respective models at that point, or $|h_l(\mathbf{x}_j)-h_p(\mathbf{x}_j)|$. Then the acquisition function at $\mathbf{x}_j$ is defined as  $\max_{l,p}(|h_l(\mathbf{x}_j)-h_p(\mathbf{x}_j)|)$, representing the maximum disagreement at $\mathbf{x}_j$ where $l,p=1,2,\ldots,Q$. Then the next data, $\mathbf{x}_{\text{QBC}}^*$ is selected such that, 

\vspace{-1em}
\begin{equation}
     \mathbf{x}_{\text{QBC}}^*=\underset{\mathbf{x}_j}{\arg\!\max}{\left(\underset{l,p}{\max}({|h_l(\mathbf{x}_j)-h_p(\mathbf{x}_j)|)}\right)}\label{QBC}
\end{equation}
\vspace{-3em}

\noindent QBC approach based on GPR allows us to exploit the regions in the search space where the function's behavior is uncertain i.e., discontinuities or change points. 
As indicated in \cite{Rasmussenbook}, the mean square (MS) continuity and differentiability of kernel functions control their flexibility. For example, let us compare GPR models with exponential, Mat\'ern $3/2$, Mat\'ern $5/2$, and squared exponential kernel functions. The non-differentiable exponential kernel generates the roughest prediction, whereas the infinitely differentiable squared exponential kernel produces the smoothest ones. An intermediate level of smoothness can be observed in Mat\'ern $3/2$ and Mat\'ern $5/2$ kernels which are one and two times MS differentiable respectively \cite{Rasmussenbook}. The sharp changes in underlying functions are captured by the rough predictions via dense queries. But the smoother prediction deviates from the underlying model at the region.
Therefore, kernel functions with different MS continuity and differentiability behave differently where the functional form is unpredictable with discontinuity or sharp change. When QBC selects the data at which committee members differ the most in their prediction, it keeps exploiting that very region.

$\bullet\quad \textbf{Maximum variance: }$
This strategy defines the acquisition function as the variance predicted by the regression model\cite{yang}. A learner's expected error can be decomposed into noise, bias, and variance \cite{cohn1996}. Since the noise is independent of the model or data, and bias is invariant given a fixed model, minimizing the variance is intuitively guaranteed to minimize the future generalization error of the model \cite{settles.survey}. Equation~\eqref{vpred} provides the Gaussian process posterior covariance, and the diagonal elements of $\text{cov}(\hat{\mathbf{f}})$ provide the predicted variance, $\mathbb{V}$. The next data is queried~following, 

\vspace{-1em}
\begin{equation}
     \mathbf{x}_{\text{VAR}}^*=\underset{\mathbf{x}_j}{\arg\!\max}\left(\mathbb{V}(\mathbf{x}_j)\right)\label{maxvar}
\end{equation}
\vspace{-3em}

\noindent Since predicted variance is higher mostly in less-explored regions, it can be implemented as an adequate exploration acquisition function to query the next data \cite{yang}. 

$\bullet\quad \textbf{Maximum entropy: }$
Another active learning strategy formulated based on uncertainty is the maximum entropy strategy \cite{settles.survey}. Entropy is an information-theoretic
measure that often has been defined as the amount of information needed to “encode” a distribution and has been related to the uncertainty of the underlying model \cite{settles.survey}. Thus minimizing model entropy can reasonably lead to revealing the model uncertainty. Shannon's entropy has been used as an acquisition function named maximum entropy sampling \cite{holub2008} or uncertainty sampling \cite{settles.survey}. Since the posterior prediction of the Gaussian process follows a multivariate normal distribution, Shannon's entropy of the distribution,
\vspace{-3em}
\begin{align*}
    H[\hat{\mathbf{f}},\text{cov}(\hat{\mathbf{f}})]=\frac{1}{2}\log|\text{cov}(\hat{\mathbf{f}})|+\frac{D}{2}\log(2\pi e)
\end{align*}
\vspace{-3em}

\noindent where $D$ is the dimension of the variable, and $\hat{\mathbf{f}}$ and cov($\hat{\mathbf{f}}$) are the predicted mean and covariance according to Eqs.~\eqref{mpred} and~\eqref{vpred} respectively \cite{Rasmussenbook}. By minimizing the learning model uncertainty, this strategy is also more prone to exploration than exploitation. The maximum entropy strategy queries the data where it has the highest entropy following, 

\vspace{-1em}
\begin{equation}
     \mathbf{x}_{\text{ENT}}^*=\underset{\mathbf{x}_j}{\arg\!\max}\left(H[\hat{\mathbf{f}},\text{cov}(\hat{\mathbf{f}})]\right)\label{maxent}
\end{equation}

\subsection{\emph{Exploration-exploitation trade-off functions}} 
As touched on briefly in the introduction, the existing approaches have tried achieving the trade-off between exploration and exploitation using different methodologies \cite{kee2018,lourencco2022new,kuleshov2014algorithms,Elreedy}. Here we have listed a few that we will compare BHEEM with.

$\bullet\quad \textbf{Static trade-off: }$ Many of the previous studies have held on to static trade-offs between exploration and exploitation \cite{kee2018}. Prior information can be used to decide on the trade-off (e.g., equal importance to both or more importance to one based on the nature of the function). Existing approaches have conducted trial and error with different trade-off values between exploration and exploitation between appropriate exploration and exploitation acquisition functions to conduct the simulated experiments, and then select the one that promises overall better accuracy \cite{lourencco2022new}. 
By fixing the exploration-exploitation trade-off throughout the whole learning process, this method reduces the computation complexity to a great extent. Nevertheless, the same trade-off cannot be expected to demonstrate similar performance for every function (as can be seen in the experimental results).

$\bullet\quad \textbf{Probabilistic trade-off: }$
Inspired by simulated annealing \cite{van1987simulated} and built on the $\epsilon-$decreasing greedy algorithm \cite{kuleshov2014algorithms}, this algorithm updates the exploration-exploitation combination in a probabilistic approach \cite{Elreedy}. For instance, the exploration probability is defined as $p_R=\alpha^{t-1}$ where $\alpha$ is less than 1 and $t$ is the current time step or iteration number. To query new data, a uniform random variable $Z$ is generated, if $Z \leq p_R$, we consider $\eta=1$, and exploration is performed, otherwise, we consider $\eta=0$, and exploitation is applied \cite{Elreedy}. Hence the strategy starts with pure exploration. The exploration probability decays gradually, and the learning model gets more prone to pure exploitation over time. Intuitively, this transition is practical since the initial queried data should focus more on exploration, and after learning the overall trend of the function, we can identify the irregular regions via exploitation. However, the transition rate depends on the choice of $\alpha$ which we fixed at $0.7$ following Elreedy et al. \cite{Elreedy}. But again, the transition rate should depend on the nature of the function and fixing it will decrease the efficiency of the learning process.

\section{Experimental results}
 In this section, we present the experimental setup and focus on the performance evaluation of the proposed methodology. We compare the performance of BHEEM with other active learning strategies over six simulated experiments and one real-world case study for predicting the property of MAX phase materials.  Later on, we discuss convergence and sensitivity analysis of the process.
 
 \subsection{\emph{Experimental setup}}
 To demonstrate the efficacy of BHEEM, we employ the Mat\'{e}rn 3/2 kernel function in the GPR model to avoid too rough (exponential kernel) or too wavy (squared exponential kernel) prediction. The kernel function is defined as, 
 
\vspace{-1em} 
\begin{equation}
    K_{\nu=3/2}(z)=\sigma_f^2\left(1+{\sqrt{3}z/l}\right)\exp\left(-{\sqrt{3}z/l}\right)\label{matern}
\end{equation}
\vspace{-3em}

\noindent where $\sigma_f^2$ is the signal variance that we fixed at $1$, $z = ||\mathbf{x}-\mathbf{x'}||_2$, and $l$ is the length-scale parameter optimized by maximizing the log marginal likelihood of the Gaussian process regression model.

While applying the QBC strategy, we maintained a committee of ten Gaussian process models, each with a different kernel function; (i) squared exponential, (ii) exponential, (iii) Mat\'{e}rn 3/2, (iv) Mat\'{e}rn 5/2, (v) rational quadratic, (vi) product of dot product and constant kernel, (vii) product of i and iii, (viii) product of i and iv, (ix) product of ii and iii, (x) product of ii and iv \cite{Rasmussenbook}. 
 For the generation of the data, we fixed the signal-to-noise (SN) ratio to 10 which we define as the decibel of the ratio of signal power to the power of the data noise.
 For the performance evaluation, we calculated the root mean squared error as,
 
 \vspace{-1em}
 \begin{equation}
     \text{RMSE} = \sqrt{\sum_{k=1}^{1000}{\frac{(\bar{f}(\mathbf{x}_k)-f(\mathbf{x}_k))^2}{1000}}}
 \end{equation}
 \vspace{-2em} 
 
\noindent where $f(\cdot)$ and $\bar{f}(\cdot)$ represent the true and predicted response respectively at the set of equidistant test points, $\{\mathbf{x}_k\}_{k=1}^{1000}$. We repeated all simulated experiments 100 times to achieve a consistent estimate of the performance. 
 
 \subsection{\emph{Simulated experiments}}
  To demonstrate the performance of BHEEM against the existing strategies, we considered the six following functions.
  \begin{itemize}
      \item 
          $F_1(x)=\begin{cases}
                  3.5\exp\left(-\frac{(x-10)^2}{200}\right)+\epsilon, & \qquad\quad\qquad\quad\quad\text{if $x\leq 25$}\\
                8-3.5\exp\left(-\frac{(x-35)^2}{200}\right)+\epsilon, & \qquad\qquad\quad\quad\quad\text{otherwise}.
              \end{cases}\qquad\qquad\;\;\quad\qquad\quad(17)$

       \item $F_2(x)=\sin(x)+2\exp(-30x^2)+\epsilon\qquad\qquad\qquad\quad\quad\quad\;\; x\in[-2,2]\qquad\qquad\quad\qquad\;\;\quad(18)$
    %
       
       \item Three-hump camel function:\\$F_3(\mathbf{x})=2x_1^2-1.05x_1^4+x_1^6/6+x_1x_2+x_2^2+\epsilon\qquad\qquad\quad x_1,x_2\in[-5,5]\qquad\qquad\qquad\;\;(19)$
    
       \item Six-hump camel function:\\$F_4(\mathbf{x})=(4-2.1x_1^2+x_1^4/3)x_1^2+x_1x_2+(4x_2^4-4)x_2^2+\epsilon\quad x_1,x_2\in[-2,2]\qquad\qquad\qquad\;(20)$

       \item Hartmann 3D function:\\
       $F_5(\mathbf{x})=-\sum_{i=1}^4\alpha_i\exp{\left(-\sum_{j=1}^3A_{ij}(x_j-P_{ij})^2\right)}+\epsilon\quad\quad x_i\in[0,1]\;\forall i=1,2,3\quad\qquad\quad(21)$\\
       
       where $\alpha=(1.0,1.2,3.0,3.2)^T$, $A=\begin{bmatrix}
           3.0 & 10 & 30\\
           0.1 & 10 & 35\\
           3.0 & 10 & 30\\
           0.1 & 10 & 35
       \end{bmatrix}$, and $P=10^{-4}\begin{bmatrix}
           3689 & 1170 & 2673\\
           4699 & 4387 & 7470\\
           1091 & 8732 & 5547\\
           381 & 5743 & 8828
       \end{bmatrix}$ 
  
        \item $F_6(\mathbf{x})=\frac{1}{2}\sum_{i=1}^{10}x_i^2-\cos\left(2\pi x_i\right)+\epsilon\quad\qquad\qquad\qquad\quad\quad x_i\in[-5,5]\;\forall i=1,2,\hdots,10\quad\;(22)$ 
        
  \end{itemize}

\begin{figure}[ht!]
      \centering
      \includegraphics[width=1\linewidth]{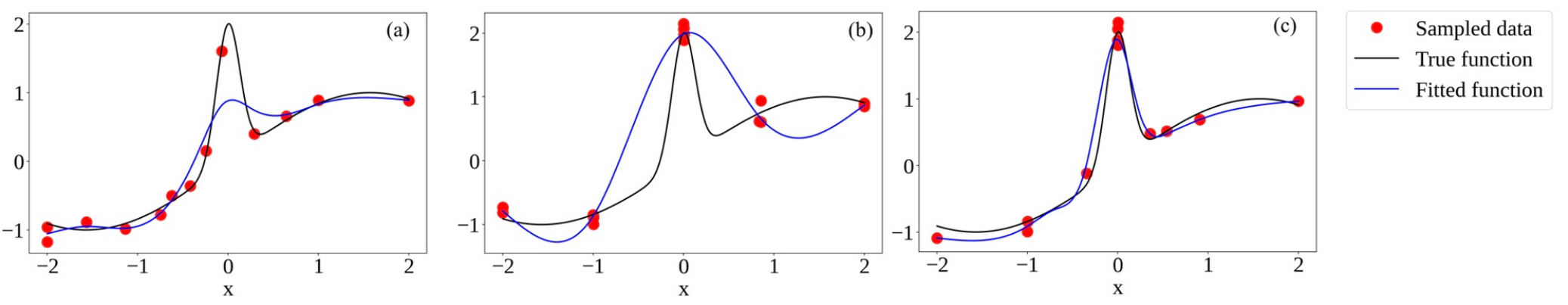}
      \caption{(a) Queried data and fitted function using exploration, (b) queried data and fitted function using exploitation, (c) queried data and fitted function using BHEEM with a dynamic trade-off between exploration and exploitation.}
      \label{toyproblem}
  \end{figure}

First, we implement BHEEM by employing iGS and QBC as the pure exploration and exploitation strategy respectively. With three random initial queried data, Figure \ref{toyproblem} shows the result of iGS, QBC, and BHEEM after the $12^\text{th}$ iteration for the underlying function from Equation~(18). Unsurprisingly, iGS spreads out its queried data in both the $x$ and $y$ direction predicting the smooth portions of the function competently, but fails to fit the peak of the function due to the sudden change near $x=0$. QBC exploits this uncertainty in predictions across different committee members who lead to different predictions near the sharp peak. Therefore, QBC queries most of the data in this region while ignoring other regions in the process. BHEEM queries data near $x=0$ enough times to fit the peak there but has also explored and provided satisfactory prediction in other regions achieving an overall lower error than the other two acquisition functions. 
  
\begin{landscape}
\begin{figure}
    \centering
    \includegraphics[width=1\linewidth]{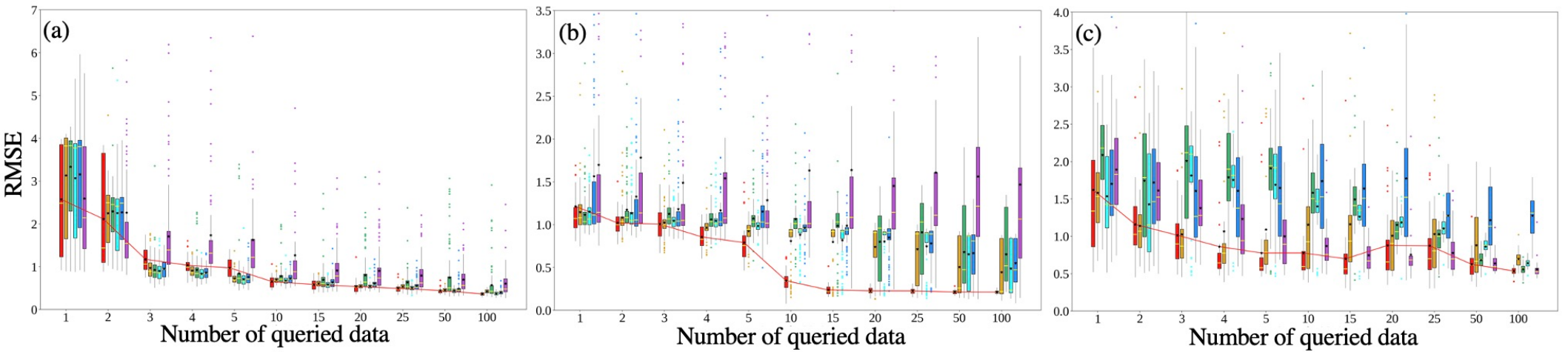}
    \includegraphics[width=1\linewidth]{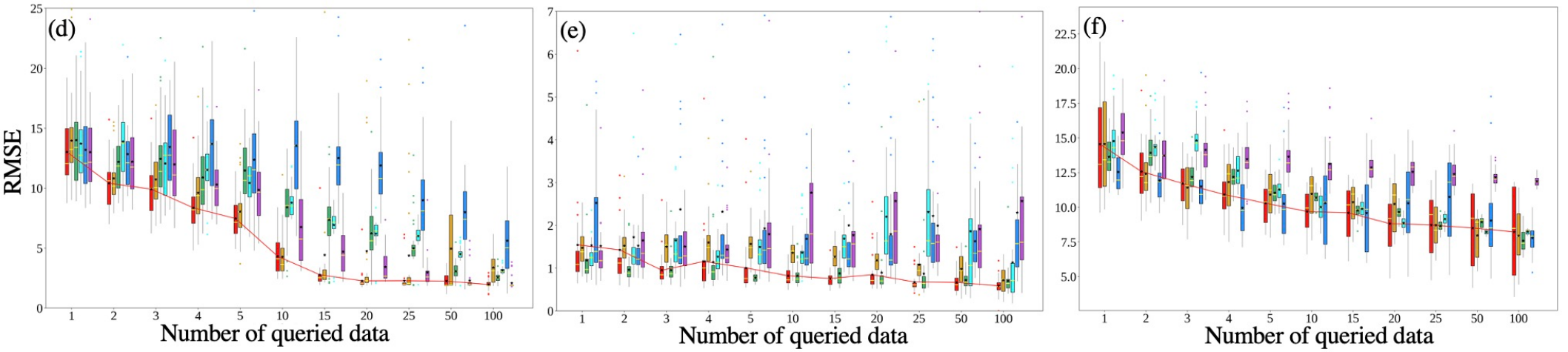}
    \includegraphics[width=1\linewidth]{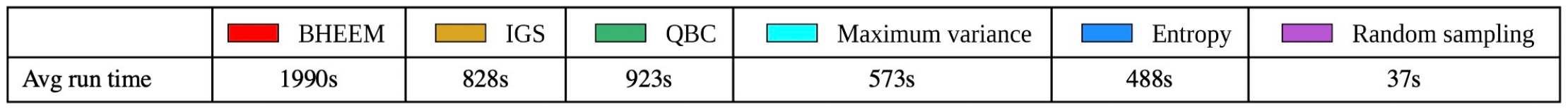}
    \caption{RMSE for BHEEM, iGS, QBC, maximum variance, maximum entropy, and random sampling strategy for (a) Equation~(17), (b) Equation~(18), (c) Equation~(19), (d) Equation~(20), (e) Equation~(21), (f) Equation~(22) as well as average computational time for all strategies.}
    \label{boxplot}
\end{figure}
\end{landscape}

Continuing to employ iGS and QBC as pure exploration and exploitation strategies respectively, Figure~\ref{boxplot}~(a-f) compares the root mean square error (RMSE) of different methodologies while learning Equation~(17-22) and summarises the result using boxplots. For each of the examples, we have set the upper limit on the total number of experiments to be 100. The key observation in the boxplots is that our proposed methodology, BHEEM, consistently achieves a lower generalization error with a smaller number of queried data than any other employed strategy. Between iGS and QBC, we observe the latter to be more efficient in some cases (Figure~\ref{boxplot}(c)), and iGS in others~(Figure~\ref{boxplot}(e)).

\noindent Nevertheless, BHEEM is either better (Figure~\ref{boxplot}(b,e)) or at least as accurate as one of the approaches (Figure~\ref{boxplot}(c)).

Among the one-dimensional problems, BHEEM achieved the lowest RMSE compared to others after the query of the first $10$ data in the case of $F_1(x)$ as observed in Figure~\ref{boxplot}(a). IGS was the closest one in this case and achieved only about $5\%$ higher RMSE on an average. While predicting $F_2(x)$, BHEEM converged the fastest and performed the best as shown in Figure~\ref{boxplot}~(b). The second~best strategy was iGS which scored about $60\%$ higher RMSE on average from BHEEM after querying $25$ data. Among the two-dimensional problems, in $F_3(\mathbf{x})$, BHEEM surpassed others most of the time and at the $50^\text{th}$ addition of point, achieved about $8.5\%$ lower RMSE from QBC which scored the closest to BHEEM for the corresponding functions. For $F_4(\mathbf{x})$, the consistency of lower generalization error and fast convergence persisted for BHEEM. For the three-dimensional $F_5(\mathbf{x})$, BHEEM and QBC were the two strategies achieving the lowest RMSE across the querying stages. BHEEM was able to achieve significantly lower RMSE than the pure exploration strategy, iGS. In the ten-dimensional problem of $F_6(\mathbf{x})$ (Figure~\ref{boxplot}(f)), all the active learning strategies are observed to compete with each other over the learning process. The pure exploration strategy, iGS, is performing better than all other acquisition functions at the $50^\text{th}$ addition of data whereas QBC seems to achieve the lowest RMSE at the $100^\text{th}$ addition. In most of these cases, all active learning strategies performed better than the random sampling strategy. Our result was affected by a few decisions including our choice of kernel, sampling algorithm and proposal distribution, number of points in the initial set of queried data, etc. However, the overall result demonstrates that our proposed approach tends to perform at least tantamount to pure exploration or pure exploitation.

To compare the computational time taken by the applied methodologies, Figure~\ref{boxplot} also presents the average time for each methodology to query 100 data during the learning process. Our proposed methodology, BHEEM, takes a significantly larger processing time, about $2.5$ and $2$ times higher than the duration of iGS and QBC, respectively. Hence, BHEEM is more suitable for offline applications that involve time-consuming physical experiments than the onlineones, which is the case for most applications in manufacturing and materials that involve conducting time consuming physical experiments.

\begin{figure}[h!]
     \centering
     \includegraphics[width=1\linewidth]{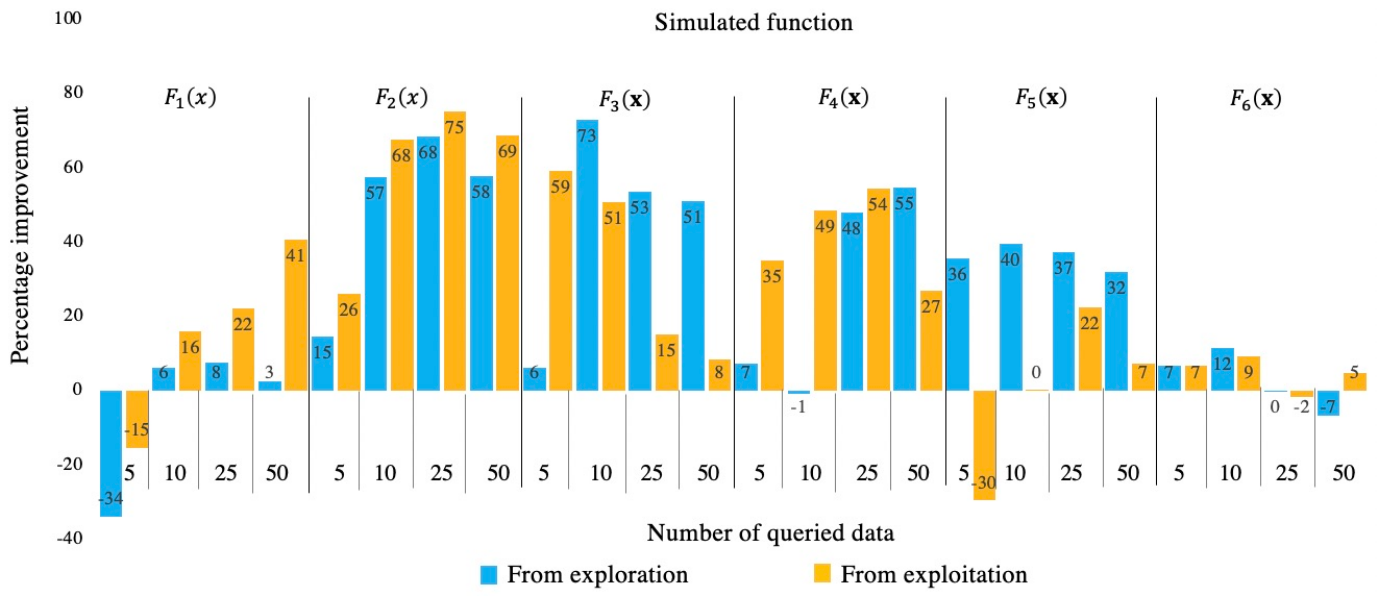}
     \caption{Percentage improvement of BHEEM from iGS and QBC acquisition function for the six simulated~studies.}
  \label{percentage} 
 \end{figure}

Figure~\ref{percentage} presents the percentage improvement of BHEEM from the pure exploration (iGS) and pure exploitation (QBC) calculated from the average RMSE obtained in the simulated examples. Corresponding to each function, the four bars of each color represent the percentage improvement from exploration using iGS (blue) and exploitation using QBC (yellow) after querying the fifth, tenth, $25^{\text{th}}$, and $50^{\text{th}}$ data. Here, the positive and negative bars indicate increased and decreased accuracy of the BHEEM respectively, compared to pure exploration and pure exploitation. We observe that among the twenty-four cases of displayed results, only one showed deterioration where BHEEM had lower accuracy than both the pure strategies. In all other stages, BHEEM had significant improvement and achieved lower RMSE compared to at least one of them while surpassing result of both pure strategies in sixteen of them.

\begin{figure}[ht!]
    \centering
    \includegraphics[width=0.9\linewidth]{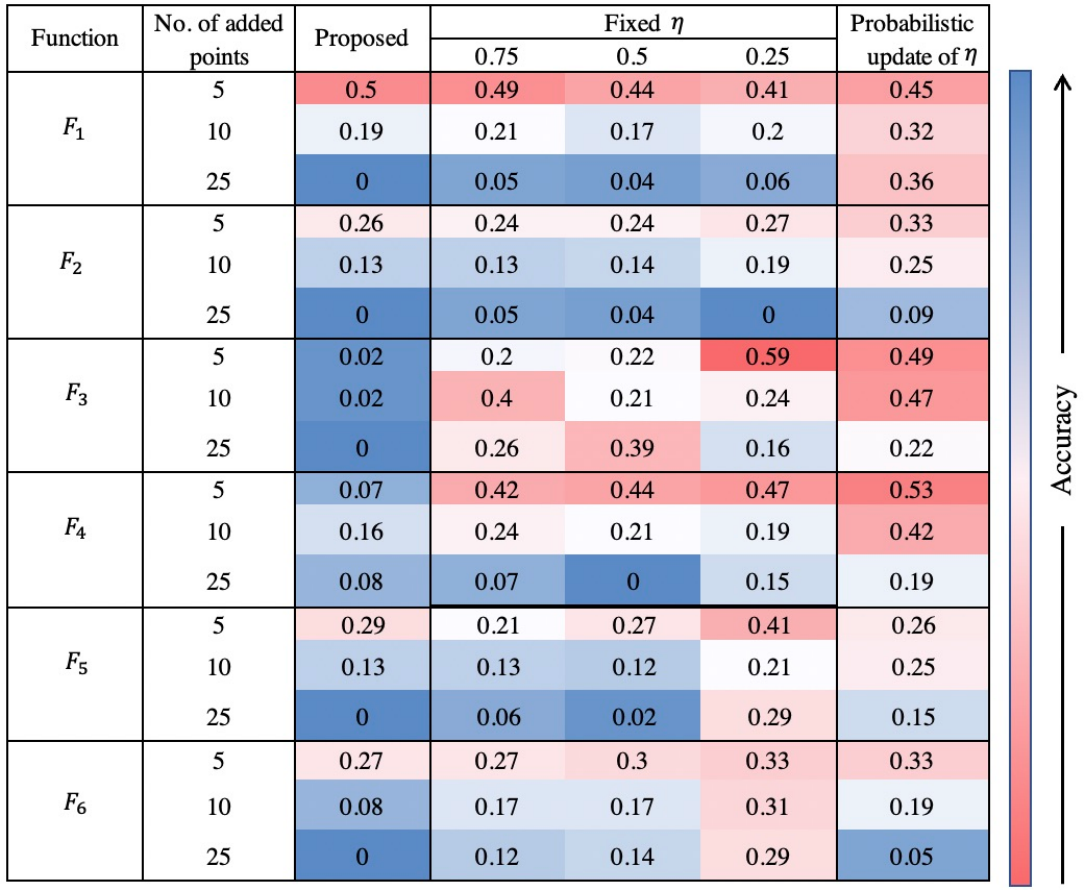}
    \caption{Average RMSE for BHEEM, static trade-off, and probabilistic trade-off after adding the 5th, 10th, and 25th data for the simulated functions.}
    \label{heatmap}
\end{figure}

We also compare our proposed methodology with static and probabilistic updates of $\eta$. For the static trade-off, we considered $\eta=0.25,0.5,0.75$ during the learning process individually and calculated the RMSE for each of them. To note, $\eta=0$ and $\eta=1$ refer to pure exploitation and pure exploration respectively which we have already compared with BHEEM in Figure~5 and Figure~6.  For the probabilistic update of $\eta$, we followed the strategy described in Section 3.3.2. A heatmap for the average RMSE scaled for each of the comparative trade-off methodologies is presented in Figure~\ref{heatmap} where blue and red cells represent more and less accurate models respectively. The heatmap clearly shows that no one value of $\eta$ works well for every function, or even for every iteration in one function. BHEEM distinctly provides better results than all the static and probabilistic trade-offs.

\begin{figure}[ht!]
    \centering
    \includegraphics[width=0.65\linewidth]{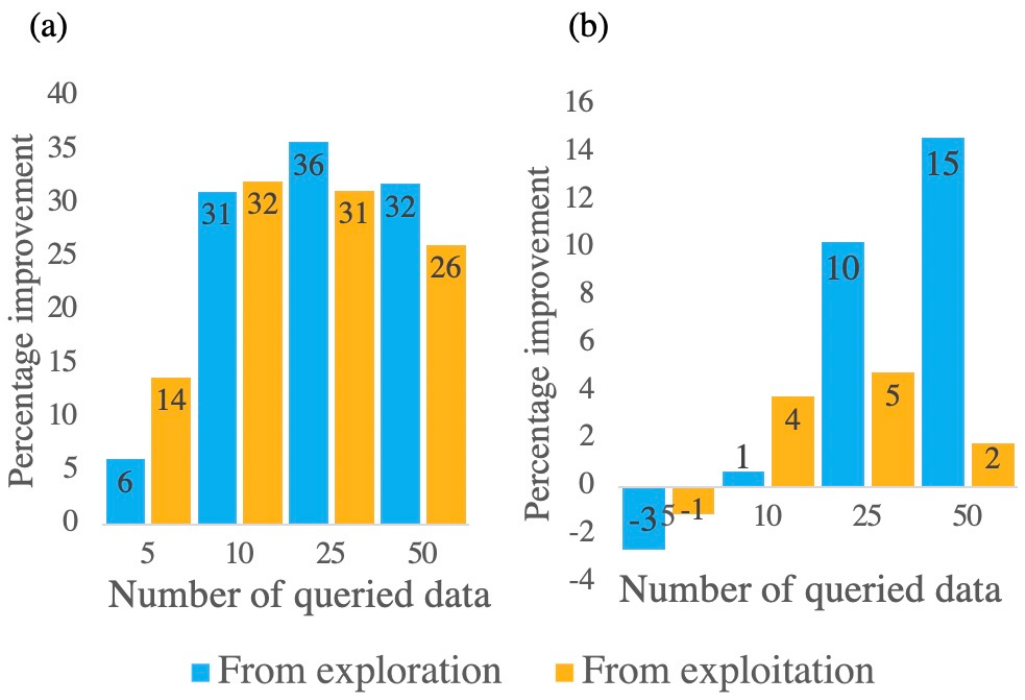}
    \caption{(a) Average percentage improvement of BHEEM from pure exploration (iGS) and pure exploitation (QBC) acquisition function across simulated studies, (b) Average percentage improvement of BHEEM from pure exploration (maximum variance) and pure exploitation (QBC) acquisition function across simulated studies.}
    \label{average}
\end{figure}

Finally, Figure~\ref{average} presents the average improvement achieved by BHEEM from pure exploration and exploitation strategy across the functions. Here, we have used iGS and maximum variance as the exploration strategy in Figure~\ref{average}(a) and Figure~\ref{average}(b) respectively, and QBC as the exploitation strategy in both cases. In Figure~\ref{average}(a), we observe about $7\%$ and $21\%$ lower average RMSE from pure exploration (iGS) and exploitation (QBC) respectively. In Figure~\ref{average}(b), we observe about $5.7\%$ and $2.3\%$ lower average RMSE from pure exploration (maximum variance) and exploitation (QBC) respectively. Overall, our proposed methodology of trade-off always promises better accuracy than pure exploration and exploitation irrespective of the choice of strategies.

 \subsection{\emph{Case study: MAX phase materials}}
 The layered ternary carbides and nitrides with the general formula $M_{n+1}AX_n$ are called MAX phase materials where $M$ and $A$ are early transition metal and A-group elements respectively, whereas $X$ is Nitrogen or Carbon and $n$ is an integer between 1 and 4 \cite{Eklund}. Their layered structures kink and delaminate the materials during deformation resulting in an unusual and unique combination of both ceramic and metallic properties which makes them attractive candidates for structural and fuel coating applications. In this study, we intended to predict the lattice constants of MAX phase materials by analyzing their compositions and elastic constants using the data from Aryal et al. \cite{aryal2014}. The lattice constant is an important piece of information to define the overall lattice structure, which helps model the microstructure evolution. To represent the discrete categorical composition features into numeric input to the models, we used one-hot encoding, a popular approach replacing the categorical variable with as many variables as categories \cite{potdar2017}. Assuming that we wish to differentiate between two materials with the same elements in M and A, and the same value of $n$, one of them is a carbide, while the other is a nitride. It is possible to represent this information with two binary variables using one-hot encoding where $0$ and $1$ represent the non-existence and existence of that element in the material respectively.

\begin{figure}[h]
     \centering
     \includegraphics[width=0.8\linewidth]{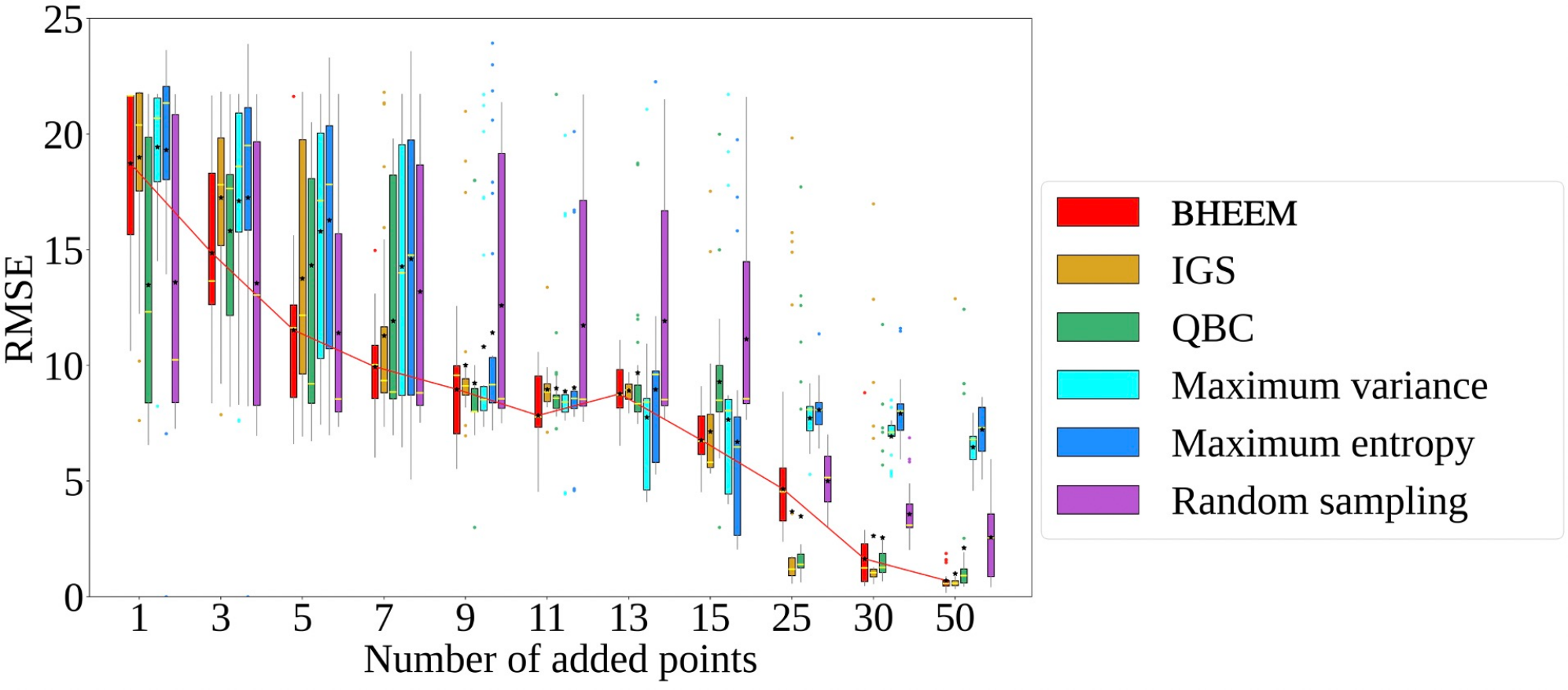}
     \caption{Comparison of RMSE for different strategies for Max phase material case study.}
  \label{fig:maxphase} 
 \end{figure}

 Each category value is represented as a 2-dimensional, sparse vector of $1$ for one of the dimensions, and $0$ for the other. In general, for variables of cardinality $d$, the one-hot encoding would transfer it to $d$ number of binary variables where each observation indicates the presence (1) or absence (0) of the dichotomous binary variable~\cite{potdar2017}. In the MAX phase problem, we have ten elements in M, twelve elements in A, and two elements in X. Hence, after using one-hot encoding and adding the numerical variable for n, we have a total of 25 variables representing the composition of the materials. Including the composition and the elastic constants, there is a total of 30 predictors to predict the lattice constant. Figure~\ref{fig:maxphase} provides a comparative analysis of RMSE of different methodologies employed in this case study. Our proposed methodology (combining iGS and QBC) outperformed the other strategies and surpassed at least one of the pure exploration or exploitation in almost every iteration. BHEEM achieved about $10.4\%$ and $11.3\%$ average improvement from the pure exploration and exploitation strategy respectively.
 
\subsection{\emph{Convergence}} 
In this section, we discuss the convergence of sampled $\eta$ at one querying stage, its relation with the choice of hyperparameters, and the dynamics $\eta$ during the active learning process. The results in this section are all based on the learning process of $F_1(x)$.

\begin{figure}[h]
    \centering
    \includegraphics[width=1\linewidth]{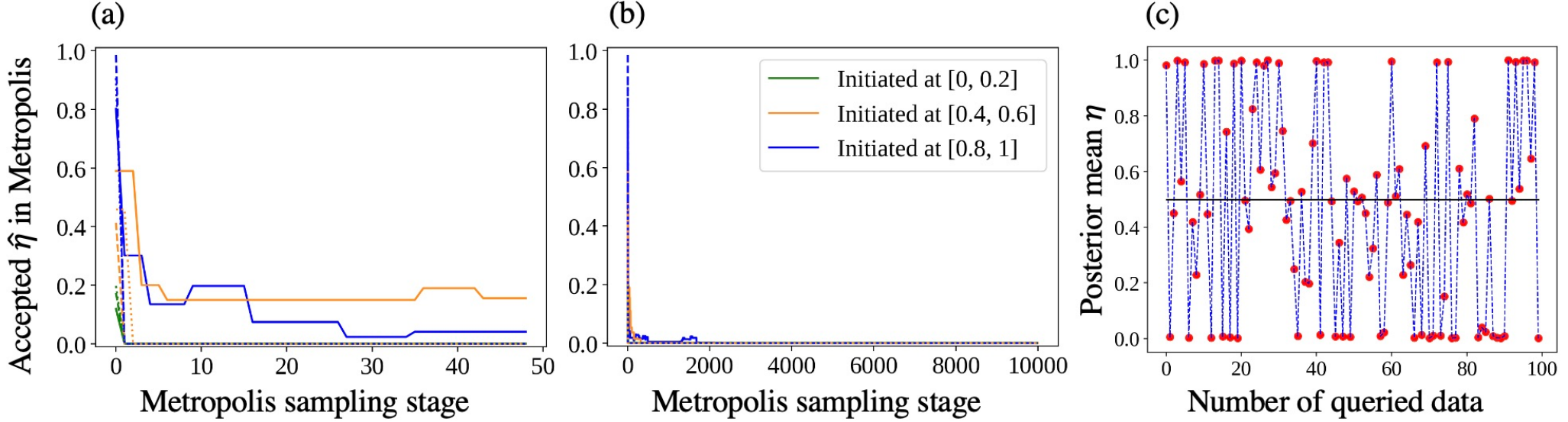}
    \caption{(a) Metropolis sampling chain with first 50 accepted $\hat{\eta}$ in one querying iteration with different initial samples, (b) Metropolis sampling chain with first 10,000 accepted $\hat{\eta}$ in one querying iteration with different initial samples. Different colored plots in each figure represent chains with initial samples at different ranges. (c) Progression of posterior mean of $\eta$ while learning a function.}
    \label{metropolis}
\end{figure}

$\bullet\quad \textbf{Convergence of Metropolis: }$ We visualize the progression of the Metropolis algorithm in Figure~\ref{metropolis} for a representative simulation example of $F_1(x)$. It plots the accepted $\hat{\eta}$ at one stage of the querying process. The chains were initiated with different ranges of values for $\hat{\eta}$ and continued till 10,000 samples were selected. The green, orange, and blue lines represent MCMC chains initiated with sampled values of $\eta$ ranging between $(0,0.2)$, $(0.4,0.6)$, and $(0.8,1)$ respectively. For each range of initial values, we plotted only three MCMC chains for better visualization. Figure~\ref{metropolis}(a) and Figure~\ref{metropolis}(b) demonstrate the chains till 50 and 10,000 accepted samples respectively. It is evident from the figures that irrespective of the initial sample, MCMC chains at one stage of learning converge to a close neighborhood of $\eta$ very quickly.

$\bullet\quad \textbf{Dynamics of $\mathbf{\eta}$: }$
Progression of the posterior mean of $\eta$ during successive query stages while learning $F_1(x)$ is presented in Figure~\ref{metropolis}(c). Dotted line is used only for visualization of the sample path and has no physical meaning. The values of $\eta$ closer to $1$ and $0$ encourages exploration and exploitation respectively. Every other value of $\eta$ combines exploration and exploitation in a specific ratio. Based on this representative sample path, we first note that BHEEM is never stuck with exploration or exploitation which is the case when the trade-off parameter is pre-defined. A close investigation of the sampled $\eta$ shows that BHEEM continues to either explore, exploit, or combine both towards optimally learning the function as more data points are queried.
Note that the sample path of the mean $\eta$ depends on the noise level therefore may be different on each run.  

\begin{figure}[h!]
        \centering
        \includegraphics[width=0.9\linewidth]{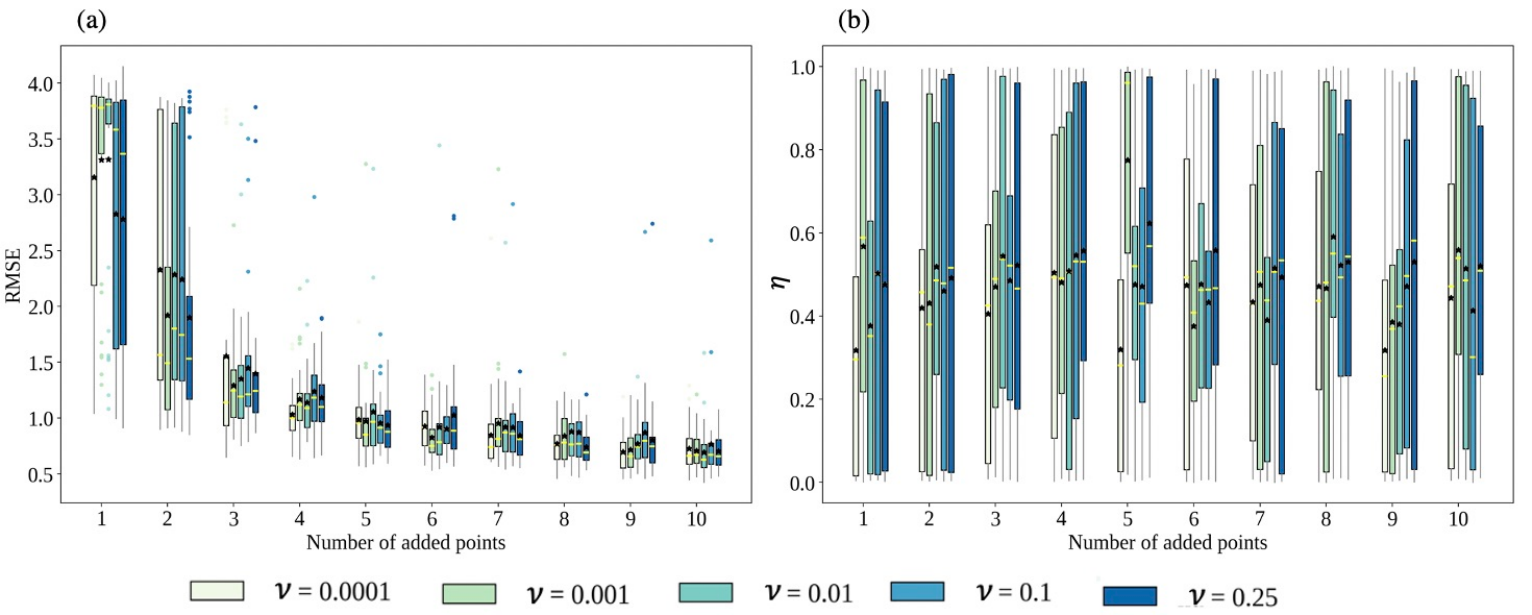}
        \includegraphics[width=0.9\linewidth]{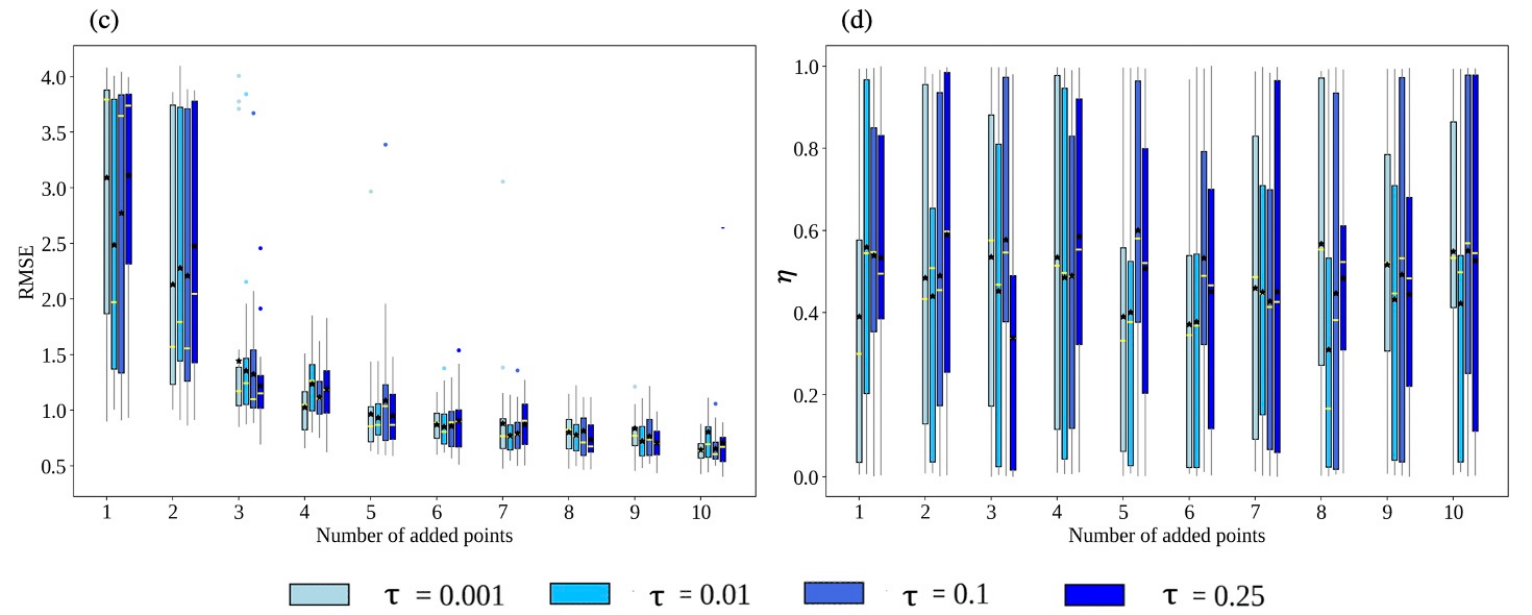}
        \caption{(a) Boxplot of root mean square error concerning $\nu$, (b) boxplot of $\eta$ with for $\nu$, (c) boxplot of root mean square error with respect to $\tau$, (d) boxplot of $\eta$ with for different $\tau$.}
        \label{tau}
\end{figure}

$\bullet\quad \textbf{Sensitivity of hyperparameters: }$
Through the sensitivity analysis of the standard deviation of proposal distribution, $\tau$, and threshold parameter, $\nu$, visualized in Figure~\ref{tau}, we will discuss the dependence of convergence of RMSE and selection of $\eta$ on both the hyperparameters. We repeated our experiments by changing $\nu$ as $0.0001,0.001,0.01,0.1,$ and $0.25$ while keeping $\tau$ fixed. Boxplot of root mean square error (RMSE) across different querying stages in Figure~\ref{tau}(a) indicates that different values of $\nu$ do not have a significant effect on RMSE the since we execute the MCMC sampling long enough to ensure convergence. However, the noise in data compels us to query the data differently each time, and the value of $\eta$ at the previous stages affect $\eta$ in subsequent stages. Therefore, selected values of $\eta$, presented as a boxplot in Figure~\ref{tau}(b), have a large variance across the learning process. Due to the near indifference to the result, we have chosen $\nu=0.001$ for all experimentation. Figure~\ref{tau}(c) and Figure~\ref{tau}(d) present the RMSE and the $\eta$ selected respectively for different values of $\tau$, $0.001,0.01,0.1,$ and $0.25$, while keeping $\nu=0.001$. Here also, we observe the behavior of RMSE and $\eta$ to be indifferent towards the value of $\tau$. But due to our previous discussion on the relation between $\tau$ and the speed of convergence of Metropolis algorithm in the last paragraph of Section 3.1.2, we chose $\tau=0.1$ throughout all experiments.

\section{Conclusion}
In this work, we address the exploration-exploitation problem in active learning for regression problems. 
Our approach, BHEEM, is based on dynamically balancing the trade-off between exploration and exploitation during the learning process using a Bayesian hierarchical model. BHEEM captures the variability in the trade-off parameter during each query stages and across successive query stages. In the absence of a likelihood function, we devised an approximate Bayesian computation based on the linear dependence of the data in the Hilbert space to sample from the posterior distribution of the trade-off parameter. 
We demonstrated and compared BHEEM with exploration, exploitation, and other well-known strategies to balance the two in the six simulated and one real-world case study. From the average percentage improvement in the simulated experiments, BHEEM achieved about $21\%$ and $24\%$ lower average RMSE from pure exploration and exploitation respectively irrespective of the chosen strategies. Overall, when compared to existing active learning approaches, BHEEM converged with fewer iterations in all cases and performed better or at least as effective as the more efficient strategy among the two it combines. The proposed approach to dynamically balancing exploration and exploitation has wide applications in various domains where conducting experiments and labeling data is costly such as materials characterization, mechanical testing, manufacturing, and medical~sciences.

Some of the limitations of the current studies include the assumption on the filtering distance threshold $\nu$ and more importantly, a restricted prior on the trade-off parameter $\eta$. In future works, we plan to consider a more expressive and flexible Dirichlet process prior to ensure consistency and faster convergence of the hierarchical model. 
 ]

\singlespacing
\small
\printbibliography

@book{Shannon,
  author    = {Claude E. Shannon}, 
  title     = {A mathematical theory of communication},
  publisher = {The Bell system technical journal},
  year      = 1948,
}

@article{pritchard1999population,
  title={Population growth of human Y chromosomes: a study of Y chromosome microsatellites.},
  author={Pritchard, Jonathan K and Seielstad, Mark T and Perez-Lezaun, Anna and Feldman, Marcus W},
  journal={Molecular Biology and Evolution},
  volume={16},
  number={12},
  pages={1791--1798},
  year={1999},
  publisher={Oxford University Press}
}

@article{engel2004kernel,
  title={The kernel recursive least-squares algorithm},
  author={Engel, Yaakov and Mannor, Shie and Meir, Ron},
  journal={IEEE Transactions on Signal Processing},
  volume={52},
  number={8},
  pages={2275--2285},
  year={2004},
  publisher={IEEE}
}

@article{marjoram2003markov,
  title={Markov chain Monte Carlo without likelihoods},
  author={Marjoram, Paul and Molitor, John and Plagnol, Vincent and Tavar{\'e}, Simon},
  journal={Proceedings of the National Academy of Sciences},
  volume={100},
  number={26},
  pages={15324--15328},
  year={2003},
  publisher={National Acad Sciences}
}

@inproceedings{busetto2009,
  title={Stable Bayesian parameter estimation for biological dynamical systems},
  author={Busetto, Alberto Giovanni and Buhmann, Joachim M},
  booktitle={2009 International Conference on Computational Science and Engineering},
  volume={1},
  pages={148--157},
  year={2009},
  organization={IEEE}
}

@article{char2018,
  title={Implementing machine learning in health care—addressing ethical challenges},
  author={Char, Danton S and Shah, Nigam H and Magnus, David},
  journal={The New England Journal of Medicine},
  volume={378},
  number={11},
  pages={981},
  year={2018},
  publisher={NIH Public Access}
}

@article{wuest2016,
  title={Machine learning in manufacturing: advantages, challenges, and applications},
  author={Wuest, Thorsten and Weimer, Daniel and Irgens, Christopher and Thoben, Klaus-Dieter},
  journal={Production \& Manufacturing Research},
  volume={4},
  number={1},
  pages={23--45},
  year={2016},
  publisher={Taylor \& Francis}
}

@article{bernardo1996,
  title={The concept of exchangeability and its applications},
  author={Bernardo, Jos{\'e} M},
  journal={Far East Journal of Mathematical Sciences},
  volume={4},
  pages={111--122},
  year={1996},
  publisher={SHIROCH REPROGRAPHICS}
}

@article{cohenshould,
  title={Should I stay or should I go? How the human brain manages},
  author={Cohen, Jonathan D and McClure, Samuel M and Angela, J Yu}
}

@article{chernoff1959,
  title={Sequential design of experiments},
  author={Chernoff, Herman},
  journal={The Annals of Mathematical Statistics},
  volume={30},
  number={3},
  pages={755--770},
  year={1959},
  publisher={JSTOR}
}

@article{angluin1988,
  title={Queries and concept learning},
  author={Angluin, Dana},
  journal={Machine Learning},
  volume={2},
  number={4},
  pages={319--342},
  year={1988},
  publisher={Springer}
}

@article{mackay1992,
  title={Information-based objective functions for active data selection},
  author={David JC MacKay},
  journal={Neural {C}omputation},
  volume={4},
  number={4},
  pages={590--604},
  year={1992},
  publisher={MIT Press One Rogers Street, Cambridge, MA 02142-1209, USA journals-info~…}
}

@inproceedings{Seung92,
	author = {Seung, H. S. and Opper, M. and Sompolinsky, H.},
	title = {Query by committee},
	booktitle = {{Proceedings of the fifth annual workshop on Computational learning theory}},
	pages = {287--294},
	year = {1992},
	month = {07},
	publisher = {Association for Computing Machinery},
	address = {New York, NY, USA},
	doi = {10.1145/130385.130417}
}

@incollection{lewis1994heterogeneous,
  title={Heterogeneous uncertainty sampling for supervised learning},
  author={Lewis, David D and Catlett, Jason},
  booktitle={Machine {L}earning {P}roceedings 1994},
  pages={148--156},
  year={1994},
  publisher={Elsevier}
}

@article{krogh95,
  title={Neural Network Ensembles, Cross Validation, and Active Learning},
  author={Anders Krogh and Jesper Vedelsby},
  journal={Advances in {N}eural {I}nformation {P}rocessing {S}ystems},
  volume={7},
  number = {7},
  year={1995}
}

@book{gelman1995,
  title={Bayesian data analysis},
  author={Gelman, Andrew and Carlin, John B and Stern, Hal S and Rubin, Donald B},
  year={1995},
  publisher={Chapman and Hall/CRC}
}

@article{guo2009bayesian,
  title={A Bayesian hierarchical model for analysis of single-nucleotide polymorphisms diversity in multilocus, multipopulation samples},
  author={Guo, Feng and Dey, Dipak K and Holsinger, Kent E},
  journal={Journal of the American Statistical Association},
  volume={104},
  number={485},
  pages={142--154},
  year={2009},
  publisher={Taylor \& Francis}
}

@article{cohn1996,
  title={Active learning with statistical models},
  author={David A Cohn and Zoubin Ghahramani and Michael I Jordan},
  journal={Journal of {A}rtificial Intelligence Research},
  volume={4},
  pages={129--145},
  year={1996}
}

@article{rabbitt1966,
  title={Errors and error correction in choice-response tasks.},
  author={Rabbitt, PM A},
  journal={Journal of Experimental Psychology},
  volume={71},
  number={2},
  pages={264},
  year={1966},
  publisher={American Psychological Association}
}

@book{sutton2018,
  title={Reinforcement learning: An introduction},
  author={Sutton, Richard S and Barto, Andrew G},
  year={2018},
  publisher={MIT press}
}

@article{Jones1998Dec,
	author = {Jones, Donald R. and Schonlau, Matthias and Welch, William J.},
	title = {{Efficient Global Optimization of Expensive Black-Box Functions}},
	journal = {Journal of Global Optimization},
	volume = {13},
	number = {4},
	pages = {455--492},
	year = {1998},
	month = {12},
	issn = {1573-2916},
	publisher = {Kluwer Academic Publishers},
	doi = {10.1023/A:1008306431147}
}

@article{baram2004,
  title={Online choice of active learning algorithms},
  author={Baram, Yoram and Yaniv, Ran El and Luz, Kobi},
  journal={Journal of Machine Learning Research},
  volume={5},
  number={Mar},
  pages={255--291},
  year={2004}
}

@book{Rasmussenbook,
  author    = {Christopher KI Williams and Carl Edward Rasmussen}, 
  title     = {Gaussian {P}rocesses for {M}achine {L}earning},
  publisher = {The MIT Press},
  year      = 2006,
}

@inproceedings{Burbidge,
  title={Active Learning for Regression Based on Query by Committee},
  author={Robert Burbidge and Jem J. Rowland and Ross D. King},
  booktitle={Lecture Notes in Computer Science},
  year={2007},
  address={Berlin, Heidelberg}
}

@article{Eklund,
         title = {Homoepitaxial growth of Ti–Si–C MAX-phase thin films on bulk {Ti3SiC2} substrates},
         author = {Per Eklund and Anand Murugaiah and Jens Emmerlich and Zs Czigàny and Jenny Frodelius and Michel W. Barsoum and Hans Högberg and Lars Hultman},
         journal = {Journal of Crystal Growth},
         volume = {304},
         number = {1},
         pages = {264--269},
         year = {2007}
}

@inproceedings{holub2008,
  title={Entropy-based active learning for object recognition},
  author={Holub, Alex and Perona, Pietro and Burl, Michael C},
  booktitle={2008 IEEE Computer Society Conference on Computer Vision and Pattern Recognition Workshops},
  pages={1--8},
  year={2008},
  organization={IEEE}
}

@article{Cebron,
  title={Active learning for object classification: from exploration to exploitation},
  author={Nicolas Cebron and Michael R. Berthold},
  journal={Data Mining and Knowledge Discovery},
  volume={18},
  pages={283--299},
  year={2009}
}

@book{hoff2009,
  title={A first course in Bayesian statistical methods},
  author={Hoff, Peter D},
  volume={580},
  year={2009},
  publisher={Springer}
}

@techreport{settles.survey,
    Author = {Burr Settles},
    Institution = {University of Wisconsin--Madison},
    Number = {1648},
    Title = {Active Learning Literature Survey},
    Type = {Computer Sciences Technical Report},
    Year = {2009}
}

@inproceedings{hu2010,
  title={EGAL: Exploration guided active learning for TCBR},
  author={Rong Hu and Sarah Jane Delany and Brian Mac Namee},
  booktitle={International Conference on Case-Based Reasoning},
  pages={156--170},
  year={2010},
  organization={Springer}
}

@inproceedings{Cai,
         title = {Maximizing expected model change for active learning in regression},
         author = {Wenbin Cai and Ya Zhang and Jun Zhou},
         booktitle = {2013 IEEE 13th International Conference on Data Mining},
         publisher ={IEEE},
         year = {2013}
}

@article{Ajdari,
        title = {An Adaptive Exploration-Exploitation Algorithm for Constructing Metamodels in Random Simulation Using a Novel Sequential Experimental Design},
        author = {Ali Ajdari and Hashem Mahlooji},
        year = {2014},
        journal = {Communications in Statistics - Simulation and Computation},
        number = {5},
        volume = {43},
        pages = {947--968}
}

@article{aryal2014,
  title={A genomic approach to the stability, elastic, and electronic properties of the MAX phases},
  author={Aryal, Sitaram and Sakidja, Ridwan and Barsoum, Michel W and Ching, Wai-Yim},
  journal={{P}hysica {S}tatus {S}olidi (b)},
  volume={251},
  number={8},
  pages={1480--1497},
  year={2014},
  publisher={Wiley Online Library}
}

@incollection{oneill2017,
  title={Model-free and model-based active learning for regression},
  author={Jack O’Neill and Sarah Jane Delany and Brian MacNamee},
  booktitle={Advances in {C}omputational {I}ntelligence {S}ystems},
  pages={375--386},
  year={2017},
  publisher={Springer}
}

@article{potdar2017,
  title={A comparative study of categorical variable encoding techniques for neural network classifiers},
  author={Potdar, Kedar and Pardawala, Taher S and Pai, Chinmay D},
  journal={International {J}ournal of {C}omputer {A}pplications},
  volume={175},
  number={4},
  pages={7--9},
  year={2017}
}

@inproceedings{yin2017,
  title={Deep similarity-based batch mode active learning with exploration-exploitation},
  author={Yin, Changchang and Qian, Buyue and Cao, Shilei and Li, Xiaoyu and Wei, Jishang and Zheng, Qinghua and Davidson, Ian},
  booktitle={2017 IEEE International Conference on Data Mining (ICDM)},
  pages={575--584},
  year={2017},
  organization={IEEE}
}

@article{kee2018,
  title={Query-by-committee improvement with diversity and density in batch active learning},
  author={Kee, Seho and Del Castillo, Enrique and Runger, George},
  journal={Information Sciences},
  volume={454},
  pages={401--418},
  year={2018},
  publisher={Elsevier}
}

@article{yang,
    title = {A variance maximization criterion for active learning},
    author = {Yazhou Yang and Marco Loog},
    journal = {Pattern Recognition}   ,
    volume = {78},
    year = {2018},
    pages = {358--370}
}

@article{Elreedy,
  title={A Novel Active Learning Regression Framework for Balancing the Exploration-Exploitation Trade-Off},
  author={Dina Elreedy and Amir F. Atiya and Samir I. Shaheen},
  journal={Entropy},
  volume={21},
  number = {7},
  year={2019}
}

@article{Wu2019,
  title={Active Learning for Regression Using Greedy Sampling},
  author={Dongrui Wu and Chin-Teng Lin and Jian Huang},
  journal={Information Sciences},
  volume={474},
  pages={90--105},
  year={2019}
}

@article{lookman,
         title={Active learning in materials science with emphasis on adaptive sampling using uncertainties for targeted design},
         author={Turab Lookman and Prasanna V. Balachandran and Dezhen Xue and Ruihao Yuan},
         year = {2019},
         journal = {npj Computational Materials},
         volume = {5},
         number = {1},
         pages = {2057--3960}
         }

@article{Liao2020Jun,
	author = {Liao, T. Warren and Li, Guoqiang},
	title = {{Metaheuristic-based inverse design of materials {\textendash} A survey}},
	journal = {Journal of Materiomics},
	volume = {6},
	number = {2},
	pages = {414--430},
	year = {2020},
	month = {6},
	issn = {2352-8478},
	publisher = {Elsevier},
	doi = {10.1016/j.jmat.2020.02.011}
}

@article{Meka,
  title={An Active Learning Methodology for Efficient Estimation of Expensive Noisy Black-Box Functions Using Gaussian Process Regression},
  author={Rajitha Meka and Adel Alaeddini and Sakiko Oyama and Kristina Langer},
  journal={IEEE Access},
  volume={8},
  pages={111460--111474},
  year={2020}
}

@incollection{van1987simulated,
  title={Simulated annealing},
  author={Van Laarhoven, Peter JM and Aarts, Emile HL},
  booktitle={Simulated annealing: Theory and applications},
  pages={7--15},
  year={1987},
  publisher={Springer}
}

@article{lourencco2022new,
  title={A new active learning approach for adsorbate--substrate structural elucidation in silico},
  author={Louren{\c{c}}o, Maicon Pierre and Herrera, Lizandra Barrios and Hosta{\v{s}}, Ji{\v{r}}{\'\i} and Calaminici, Patrizia and K{\"o}ster, Andreas M and Tchagang, Alain and Salahub, Dennis R},
  journal={Journal of Molecular Modeling},
  volume={28},
  number={6},
  pages={1--11},
  year={2022},
  publisher={Springer}
}

@inproceedings{afifi2020,
  title={Reinforcement Learning for Virtual Network Embedding in Wireless Sensor Networks},
  author={Afifi, Haitham and Karl, Holger},
  booktitle={2020 16th International Conference on Wireless and Mobile Computing, Networking and Communications (WiMob)(50308)},
  pages={123--128},
  year={2020},
  organization={IEEE}
}

@misc{Chen2021,
    author    = {Zhehui Chen and Simon Mak and C. F. Jeff Wu},
    title     = {A hierarchical expected improvement method for Bayesian optimization},
    url       = "arXiv:1911.07285"
}

@inproceedings{loy2012stream,
  title={Stream-based joint exploration-exploitation active learning},
  author={Loy, Chen Change and Hospedales, Timothy M and Xiang, Tao and Gong, Shaogang},
  booktitle={2012 IEEE Conference on Computer Vision and Pattern Recognition},
  pages={1560--1567},
  year={2012},
  organization={IEEE}
}

@inproceedings{dasgupta2008hierarchical,
  title={Hierarchical sampling for active learning},
  author={Dasgupta, Sanjoy and Hsu, Daniel},
  booktitle={Proceedings of the 25th International Conference on Machine Learning},
  pages={208--215},
  year={2008}
}

@inproceedings{li2013adaptive,
  title={Adaptive active learning for image classification},
  author={Li, Xin and Guo, Yuhong},
  booktitle={Proceedings of the IEEE Conference on Computer Vision and Pattern Recognition},
  pages={859--866},
  year={2013}
}

@article{smith2020imprecise,
  title={Imprecise action selection in substance use disorder: Evidence for active learning impairments when solving the explore-exploit dilemma},
  author={Smith, Ryan and Schwartenbeck, Philipp and Stewart, Jennifer L and Kuplicki, Rayus and Ekhtiari, Hamed and Paulus, Martin P and Tulsa 1000 Investigators and others},
  journal={Drug and alcohol dependence},
  volume={215},
  pages={108208},
  year={2020},
  publisher={Elsevier}
}

@article{koulouriotis2008reinforcement,
  title={Reinforcement learning and evolutionary algorithms for non-stationary multi-armed bandit problems},
  author={Koulouriotis, Dimitris E and Xanthopoulos, A},
  journal={Applied Mathematics and Computation},
  volume={196},
  number={2},
  pages={913--922},
  year={2008},
  publisher={Elsevier}
}

@article{kuleshov2014algorithms,
  title={Algorithms for multi-armed bandit problems},
  author={Kuleshov, Volodymyr and Precup, Doina},
  journal={arXiv preprint arXiv:1402.6028},
  year={2014}
}

@article{auer2002using,
  title={Using confidence bounds for exploitation-exploration trade-offs},
  author={Auer, Peter},
  journal={Journal of Machine Learning Research},
  volume={3},
  number={Nov},
  pages={397--422},
  year={2002}
}

@article{ishii2002control,
  title={Control of exploitation--exploration meta-parameter in reinforcement learning},
  author={Ishii, Shin and Yoshida, Wako and Yoshimoto, Junichiro},
  journal={Neural Networks},
  volume={15},
  number={4-6},
  pages={665--687},
  year={2002},
  publisher={Elsevier}
}

@inproceedings{beluch2018power,
  title={The power of ensembles for active learning in image classification},
  author={Beluch, William H and Genewein, Tim and N{\"u}rnberger, Andreas and K{\"o}hler, Jan M},
  booktitle={Proceedings of the IEEE Conference on Computer Vision and Pattern Recognition},
  pages={9368--9377},
  year={2018}
}

@article{lu2021strategies,
  title={Strategies for sequential design of experiments and augmentation},
  author={Lu, Lu and Anderson-Cook, Christine M},
  journal={Quality and Reliability Engineering International},
  volume={37},
  number={5},
  pages={1740--1757},
  year={2021},
  publisher={Wiley Online Library}
}

@article{chen2017sequential,
  title={Sequential designs based on Bayesian uncertainty quantification in sparse representation surrogate modeling},
  author={Chen, Ray-Bing and Wang, Weichung and Wu, CF Jeff},
  journal={Technometrics},
  volume={59},
  number={2},
  pages={139--152},
  year={2017},
  publisher={Taylor \& Francis}
}

@article{joseph2015sequential,
  title={Sequential exploration of complex surfaces using minimum energy designs},
  author={Joseph, V Roshan and Dasgupta, Tirthankar and Tuo, Rui and Wu, CF Jeff},
  journal={Technometrics},
  volume={57},
  number={1},
  pages={64--74},
  year={2015},
  publisher={Taylor \& Francis}
}

@article{mercer1909xvi,
  title={Xvi. functions of positive and negative type, and their connection the theory of integral equations},
  author={Mercer, James},
  journal={Philosophical Transactions of the Royal Society of London. Series A, containing papers of a mathematical or physical character},
  volume={209},
  number={441-458},
  pages={415--446},
  year={1909},
  publisher={The Royal Society London}
}
\end{document}